\documentclass[10pt,twocolumn,letterpaper]{article}

\usepackage{cvpr}
\usepackage{times}
\usepackage{epsfig}
\usepackage{graphicx}
\usepackage{amsmath}
\usepackage{amssymb}
\usepackage{bbm}
\usepackage{bm}
\usepackage{algorithm}
\usepackage[noend]{algpseudocode}
\usepackage{mathrsfs}
\newcounter{alphasect}
\usepackage{enumitem}

\def\alphainsection{0}

\let\oldsection=\section
\def\section{%
  \ifnum\alphainsection=1%
    \addtocounter{alphasect}{1}
  \fi%
\oldsection}%
\renewcommand\thesection{%
 \ifnum\alphainsection=1%
   \Alph{alphasect}%
 \else
   \arabic{section}%
 \fi%
}%

\newenvironment{alphasection}{%
  \ifnum\alphainsection=1%
    \errhelp={Let other blocks end at the beginning of the next block.}
    \errmessage{Nested Alpha section not allowed}
  \fi%
  \setcounter{alphasect}{0}
  \def\alphainsection{1}
}{%
  \setcounter{alphasect}{0}
  \def\alphainsection{0}
}%

\usepackage[pagebackref=true,breaklinks=true,letterpaper=true,colorlinks,bookmarks=false]{hyperref}

\cvprfinalcopy 

\def\cvprPaperID{10031} 

\ifcvprfinal\fi
\begin{document}

\title{Bodies at Rest: 3D Human Pose and Shape Estimation \\ from a Pressure Image using Synthetic Data\vspace{-1mm}}

\author{\hspace{-3mm}
\scalebox{0.985}{Henry M. Clever$^1$, Zackory Erickson$^1$, Ariel Kapusta$^1$, Greg Turk$^1$, C. Karen Liu$^2$, and Charles C. Kemp$^1$}
\vspace{5mm}\\
\scalebox{0.8}{$^1$Georgia Institute of Technology, Atlanta, GA, USA, $^2$Stanford University, Stanford, CA, USA}
\vspace{-1mm}\\
\scalebox{0.78}{{\tt\small \hspace{-5mm}\{henryclever, zackory, akapusta\}@gatech.edu, turk@cc.gatech.edu, karenliu@cs.stanford.edu, charlie.kemp@bme.gatech.edu}}
}

\twocolumn[{%
\renewcommand\twocolumn[1][]{#1}%
\maketitle

\begin{center}
\centering
\vspace{-0.2cm}
\includegraphics[width=17.5cm]{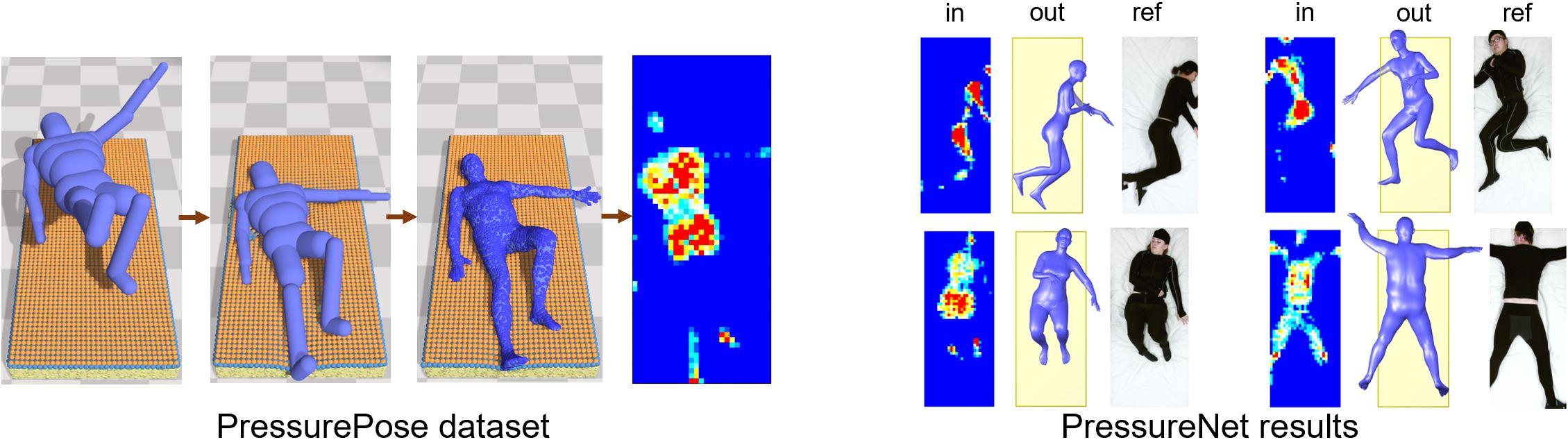}
\end{center}
\vspace{-0.2cm}

Figure 1. Left: The PressurePose dataset has 206K 3D human poses and shapes with pressure images generated by physics simulations that drop articulated rigid body models and soft body models on a soft body model of a bed and pressure sensing mat. Right: PressureNet is a deep learning model trained on synthetic data that performs well on real data: pressure image input with gender (in), 3D human mesh output (out), RGB image for reference (ref).

\vspace{6mm}
}]

\setcounter{figure}{1}  

\begin{abstract}
\vspace{-1mm}
People spend a substantial part of their lives at rest in bed. 3D human pose and shape estimation for this activity would have numerous beneficial applications, yet line-of-sight perception is complicated by occlusion from bedding. Pressure sensing mats are a promising alternative, but training data is challenging to collect at scale. We describe a physics-based method that simulates human bodies at rest in a bed with a pressure sensing mat, and present PressurePose, a synthetic dataset with 206K pressure images with 3D human poses and shapes. We also present PressureNet, a deep learning model that estimates human pose and shape given a pressure image and gender. PressureNet incorporates a pressure map reconstruction (PMR) network that models pressure image generation to promote consistency between estimated 3D body models and pressure image input. In our evaluations, PressureNet performed well with real data from participants in diverse poses, even though it had only been trained with synthetic data. When we ablated the PMR network, performance dropped substantially. 



\end{abstract}

\setlength{\abovedisplayskip}{-7pt}
\setlength{\belowdisplayskip}{7pt}

\pagestyle{plain}
\vspace{-4mm}
\section{Introduction}
Humans spend a large part of their lives resting. While resting, humans select poses that can be sustained with little physical exertion. Our primary insight is that human bodies at rest can be modeled sufficiently well to generate synthetic data for machine learning. The lack of physical exertion and absence of motion makes this class of human activities amenable to relatively simple biomechanical models similar to the ragdoll models used in video games \cite{millington2010game}.


We apply this insight to the problem of using a pressure image to estimate the 3D human pose and shape of a person resting in bed. This capability would be useful for a variety of healthcare applications such as bed sore management \cite{farshbaf2013detecting}, tomographic patient imaging \cite{grimm2012markerless}, sleep studies \cite{casas2019patient}, patient monitoring \cite{chen2018patient}, and assistive robotics \cite{clever20183d}. To this end, we present the PressurePose dataset, a large-scale synthetic dataset consisting of 3D human body poses and shapes with pressure images (Fig. 1, left). We also present PressureNet, a deep learning model that estimates 3D human body pose and shape from a low-resolution pressure image (Fig. 1, right).  

Prior work on the problem of human pose estimation from pressure images \cite{casas2019patient, clever20183d, grimm2012markerless, harada2001pressure,  liu2014bodypart} has primarily used real data that is challenging to collect. Our PressurePose dataset has an unprecedented diversity of body shapes, joint angles, and postures with more thorough and precise annotations than previous datasets (Table \ref{tab:literature_table}). While recent prior work has estimated 3D human pose from pressure images, \cite{casas2019patient,clever20183d}, to the best of our knowledge PressureNet is the first system to also estimate 3D body shape.

Our synthetic data generation method first generates diverse samples from an 85 dimensional human pose and shape space. After rejecting samples based on self-collisions and Cartesian constraints, our method uses each remaining sample to define the initial conditions for a series of two physics simulations. The first finds a body pose that is at rest on a simulated bed. Given this pose, the second physics simulation generates a synthetic pressure image. 

Our method uses SMPL \cite{loper2015smpl} to generate human mesh models and a capsulized approximation of SMPL \cite{bogo2016keep} to generate articulated rigid-body models. The first physics simulation drops a capsulized articulated rigid-body model with low-stiffness, damped joints on a soft-body model of a bed and pressure-sensing mat. Once the articulated body has settled into a statically stable configuration, our method converts the settled capsulized model into a particle-based soft body without articulation. This soft body model represents the shape of the body, which is important for pressure image synthesis. The second physics simulation drops this soft-body model from a small height onto the soft-body bed and sensor model. Once settled, the simulated sensor produces a pressure image, which is stored along with the settled body parameters. 

Our deep learning model, PressureNet, uses a series of two networks modules. Each consists of a convolutional neural network (CNN) based on \cite{clever20183d}, a kinematic embedding model from \cite{kanazawa2018end} that produces a SMPL mesh \cite{loper2015smpl}, and a pressure map reconstruction (PMR) network. The PMR network serves as a model of pressure image generation. It is a novel component that encourages consistency between the mesh model and the pressure image input. Without it, we found that our deep learning models would often make mistakes that neglected the role of contact between the body and the bed, such as placing the heel of a foot at a location some distance away from an isolated high pressure region. 

When given a mesh model of the human body, the PMR network outputs an approximate pressure image that the network can more directly compare to the pressure image input. These approximate pressure images are used in the loss function and as input to a second residual network trained after the first network to correct these types of errors and generally improve performance. 

In our evaluation, we used a commercially available pressure sensing mat (BodiTrak BT-3510~\cite{boditrak}) placed under the fitted sheet of an Invacare Homecare Bed~\cite{invacare}. This sensing method has potential advantages to line-of-sight sensors due to occlusion of the body from bedding and other sources, such as medical equipment. However, the mat we used provides low-resolution pressure images (64$\times$27) with limited sensitivity and dynamic range that make the estimation problem more challenging.

We only trained PressureNet using synthetic data, yet it performed well in our evaluation with real data from 20 people, including successfully estimating poses that have not previously been reported in the literature, such as supine poses with hands behind the head. To improve the performance of the model with real data, we used custom calibration objects and an optimization procedure to match the physics simulation to the real world prior to synthesizing the training data. We also created a noise model in order to apply noise to the synthetic pressure images when training PressureNet.

Our contributions include the following:

\begin{itemize}[noitemsep,nolistsep]

\item A physics-based method to generate simulated human bodies at rest and produce synthetic pressure images.

\item The PressurePose dataset, which consists of (1) 206K synthetic pressure images (184K train / 22K test) with associated 3D human poses and shapes~\footnote{\scriptsize{Synthetic dataset:   \texttt{doi.org/10.7910/DVN/IAPI0X}}} and (2) 1,051 real pressure images and RGB-D images from 20 human participants~\footnote{\scriptsize{Real dataset:    \texttt{doi.org/10.7910/DVN/KOA4ML}}}.

\item PressureNet~\footnote{\scriptsize{Code: \texttt{github.com/Healthcare-Robotics/bodies-at-rest}}}, a deep learning model trained on synthetic data that estimates 3D human pose and shape given a pressure image and gender.

\end{itemize}

\begin{figure*}
\begin{center}
\includegraphics[width=\textwidth]{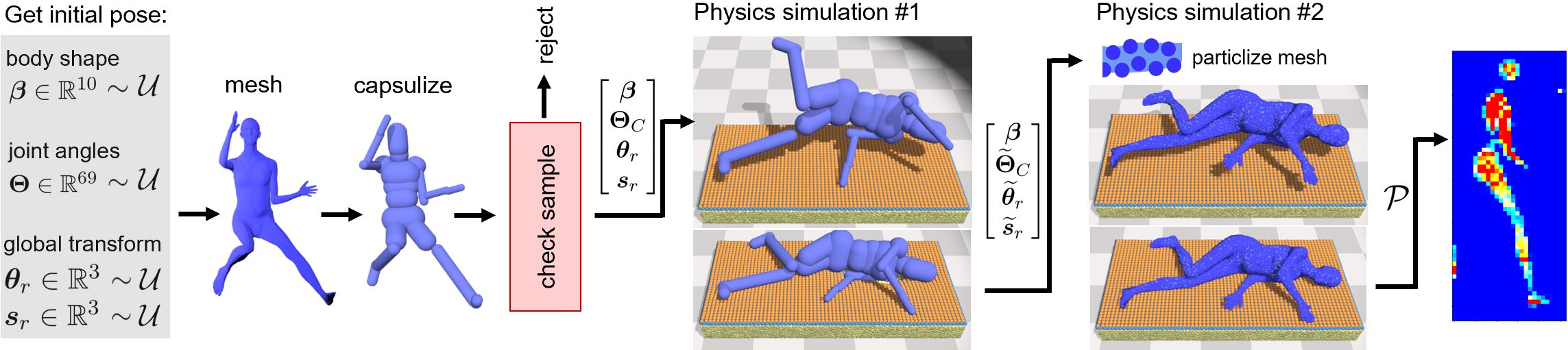}
\end{center}
\vspace{-4.5mm}
    \caption{We generate the initial pose from scratch, using random sampling of the body shape, joint angles, and global transform on the bed. We use rejection sampling to distribute the poses and remove self-collisions. Then, we rest a dynamic capsulized human model onto a soft bed using DartFleX, a fusion of DART and FleX simulators, to get an updated resting pose. Because this model is a rather rough approximation of human shape, we then use FleX to particlize a finer body representation to get the pressure image. }
\vspace{-3mm}
\label{fig:pipeline}
\end{figure*}

\section{Related work}\label{sec:lit_review}


\begin{table}
\begin{center}
\footnotesize

\renewcommand{\arraystretch}{1.1}
\vspace{1mm}

\scalebox{0.85}{\begin{tabular} {c|c|c|c|c|c|c|c|c}
 work & 
 \parbox[t]{1mm}{{\rotatebox[origin=c]{90}{ data: (R)eal, (S)ynth}}} & 
 \parbox[t]{3mm}{{\rotatebox[origin=c]{90}{ modality:}}} 
 \parbox[t]{3mm}{{\rotatebox[origin=c]{90}{ (P)ressure, (D)epth}}}
 \parbox[t]{3mm}{{\rotatebox[origin=c]{90}{ (T)thermal, IRS - }}} 
 \parbox[t]{1mm}{{\rotatebox[origin=c]{90}{ infrared selective}}} &  
 \parbox[t]{2mm}{{\rotatebox[origin=c]{90}{ 3D: (Y)es, (N)o}}}& 
 \parbox[t]{2mm}{{\rotatebox[origin=c]{90}{human }}}
 \parbox[t]{3mm}{{\rotatebox[origin=c]{90}{representation:}}}
 \parbox[t]{1mm}{{\rotatebox[origin=c]{90}{(S)keleson, (M)esh}}}& 
 \parbox[t]{2mm}{{\rotatebox[origin=c]{90}{ postures}}} & 
 \parbox[t]{2mm}{{\rotatebox[origin=c]{90}{ \# joints}}}&
 \parbox[t]{2mm}{{\rotatebox[origin=c]{90}{ \# identities}}}& 
 \parbox[t]{2mm}{{\rotatebox[origin=c]{90}{ \# images}}} \\
\hline
\hline
\cite{harada2001pressure} & R & P & Y & M & SP+, K & 18 & 1 & ? \\
\hline
\cite{grimm2012markerless} & R & D, P & N & S & SP, L, P & 10 & 16 & 1.1 K \\
\hline
\cite{liu2014bodypart} & R & P& N & S & SP, L & 8* & 12 & 1.4 K \\
\hline
\cite{achilles2016patient} & R & D & Y & S & I/O, SP, L & 14 & 10 & 180 K \\
\hline
\cite{chen2018patient} & R & RGB & N & S & SP, UNK & 7 & 3 & 13 K \\
\hline
\cite{clever20183d}& R & P & Y & S & SP, ST, K & 14 & 17 & 28 K\\
\hline
\cite{casas2019patient} & R & P & Y & S & SP+, L+, & 14 & 6 & 60 \\
& & & & & ST & & & \\
\hline
\cite{liu2019deep} & R & IRS & N & S & SP+, L+ & 14 & 2  & 419 \\
\hline
\cite{liu2019seeing} & R & T & N & S &  SP+, L+ & 14 & 109 & 14 K \\
\hline
Ours & S/ & P & Y & M & SP+, L+, & 24 & 200K/ & 200K/\\
& R & & & &  P+, K, CL & & 20 & 1K \\
& & & & & HBH, PHU & & & \\
\hline
\cline{1-9}
\multicolumn{9}{l}{posture key: SP - supine. L - lateral. P - prone. K - knee raised. I/O - getting in/out}\\
\multicolumn{9}{l}{of bed. ST - sitting. CL - crossed legs. HBH - hands behind head. PHU - prone}\\
\multicolumn{9}{l}{hands up. + indicates a continuum between postures. * indicates limbs.}\\

\end{tabular}}
\label{tab:literature_table}
\label{table:error}
\end{center}
\vspace{-0.4cm}
\caption{Comparison of Literature: Human Pose in Bed.}\label{tbl:seated}
\vspace{-0.3cm}
\end{table}

\textbf{Human pose estimation.}
There is long history of human pose estimation from camera images \cite{agarwal2006recovering,liu2014bodypart,okada2008relevant,sarafianos20163dhuman, shotton2011real} and the more recent use of CNNs \cite{tompson2014joint, toshev2014deep}. The field has been moving rapidly with the estimation of 3D skeleton models \cite{pavlakos2017coarse, zhou2016deep}, and human pose and shape estimation as a 3D mesh \cite{bogo2016keep, kanazawa2018end, pavlakos2019expressive} using human body models such as SCAPE \cite{anguelov2005scape} and SMPL \cite{loper2015smpl}. These latter methods enforce physical constraints to provide kinematically feasible pose estimates, some via optimization \cite{bogo2016keep} and others using learned embedded kinematics models \cite{clever20183d, kanazawa2018end, zhou2016deep}. Our approach builds on these works both directly through the use of available neural networks (e.g, SMPL embedding) and conceptually.

While pressure image formation differs from conventional cameras, the images are visually interpretable and methods developed in the vision community are well suited to pressure imagery~\cite{carreira2016human, kanazawa2018end, toshev2014deep}. PressureNet's model of pressure image generation relates to recent work on physical contact between people and objects \cite{brahmbhatt2019contactdb, hassan2019resolving, hasson2019learning}. It also relates to approaches that fine-tune estimates based on spatial differences between maps at distinct stages of estimation \cite{bulat2016human, carreira2016human, newell2016stacked,toshev2014deep}. 

\textbf{Human pose at rest.}
Human pose estimation has tended to focus on active poses. 
Poses in bed have attracted special attention due to their relevance to healthcare. Table \ref{tab:literature_table} provides an overview of work on the estimation of human pose for people in bed. These efforts have used a variety of sensors including RGB cameras \cite{chen2018patient}, infrared lighting and cameras for darkened rooms \cite{liu2019deep}, depth cameras to estimate pose underneath a blanket profile \cite{achilles2016patient},  thermal cameras to see through a blanket \cite{liu2019seeing}, and pressure mats underneath a person \cite{casas2019patient, clever20183d, davoodnia2019bed, grimm2012markerless,harada2001pressure, liu2014bodypart}. 

Researchers have investigated posture classification for people in bed \cite{farshbaf2013detecting, grimm2012markerless, ostadabbas2014bed}. There has been a lack of consensus on body poses to consider, as illustrated by Table \ref{tab:literature_table}. Some works focus on task-related poses, such as eating \cite{achilles2016patient}, and stretching \cite{casas2019patient}. Poses can increase ambiguity for particular modalities, such as lack of contact on a pressure mat (e.g. knee in the air) \cite{clever20183d, harada1999body} or overlapping body parts facing a thermal camera \cite{liu2019seeing}.  

Large datasets would be valuable for deep learning and evaluation. While some bed pose work has used thousands of images they have either had few participants \cite{chen2018patient} or poses highly concentrated in some areas due to many frames being captured when there is little motion \cite{achilles2016patient, casas2019patient, clever20183d}. An exception is recent work by Liu et al. \cite{liu2019seeing}, which has 109 participants. 

\textbf{Generating data in simulation.}
Approaches for generating synthetic data that model humans in the context of deep learning include physics-based simulators such as DART \cite{lee2018dart} and PyBullet \cite{erickson2019assistivegym} and position-based dynamics simulators such as PhysX \cite{erickson2018deep} and FleX \cite{macklin2014unified}. Some have used these tools to simulate deformable objects like cloth \cite{clegg2017learning,   erickson2018deep}. For vision, creating synthetic depth images is relatively straightforward (e.g.~\cite{achilles2016patient}) while RGB image synthesis relies on more complex graphics approaches \cite{chen2016synth, varol2017learning, yu2019simulcap}.  

Some past works have simulated pressure sensors.  One approach is to model the array as a deformable volume that penetrates the sensed object, where force is a function of distance penetrated~\cite{pezzementi2010characterization}. Others model pressure sensing skin as a mass-spring-damper array~\cite{habib2014skinsim, hollis2018bubble}; the former considers separate layers for the skin and the sensor, a key attribute of pressure arrays covering deformable objects. 

\section{PressurePose Dataset Generation}\label{sec:methods1}
Our data generation process consists of three main stages, as depicted in Fig.~\ref{fig:pipeline}: sampling of the body pose and shape; a physics simulation to find a body pose at rest; and a physics simulation to generate a pressure image. We use two simulation tools, FleX (Section~\ref{ssec:flex}) for simulating soft body dynamics, and DART (Section~\ref{ssec:dartflex}) for articulated rigid body dynamics.

\begin{figure*}
\begin{center}
\includegraphics[width=\textwidth]{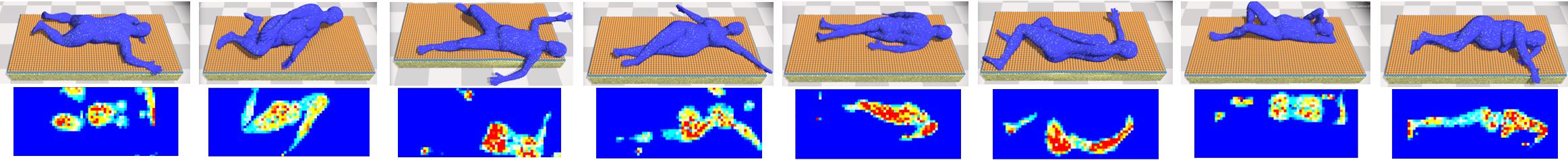}
\end{center}
\vspace{-5mm}
    \caption{Physics simulation \#2 output: PressurePose synthetic dataset examples.}
\vspace{-2mm}
\label{fig:pressurepose}
\vspace{-3mm}
\end{figure*}


\textbf{Sample initial pose and shape.} 
We sample initial pose (i.e. joint angles) and body shape parameters from the SMPL human model~\cite{loper2015smpl}. The pose consists of 69 joint angles, $\boldsymbol{\Theta} \in \mathbbm{R}^{69}$,  
which we sample from a uniform distribution, $\mathcal{U}$, bounded by joint angle limits defined for the hips, knees, shoulders, and elbows in~\cite{boone1979normal, roaas1982normal, soucie2011range}. We initialize the human body above the bed with a uniformly sampled roll $\theta_{r,1}$, yaw $\theta_{r,3}$, and 2D translation $\{ s_{r,1}, s_{r,2} \}$ across the surface of the bed. The pitch $\theta_{r,2}$ is set to $0$ and the distance normal to the bed $s_{r,3}$ is based on the position of the lowest initial joint position. This determines the global transform, $\{\boldsymbol{\theta}_r$, $\boldsymbol{s}_r \} \in \mathbbm{R}^6$. The shape of a SMPL human is determined from a set of 10 PCA parameters, $\boldsymbol{\beta} \in \mathbbm{R}^{10}$, which we also sample uniformly, bounded by $[-3,3]$ following~\cite{ranjan2018learning}. We use rejection sampling in three ways for generating initial poses: to more uniformly distribute overall pose about the Cartesian space (rather than the uniformly sampled joint space), to create a variety of data partitions representing specific common postures (e.g. hands behind the head), and to reject pose samples when there are self-collisions. See Appendix A.1. 
This step outputs pose and shape parameters
$\{ \bm{\beta}, \boldsymbol{\Theta}_C,  \bm{\theta}_r, \bm{s}_r \}$, where $\boldsymbol{\Theta}_C$ is a set of joint angles conditioned on $\bm{\beta}$ that has passed these criteria. 


\textbf{Physics Simulation \#1: Resting Pose.} 
We use FleX \cite{macklin2014unified} to simulate a human model resting on a soft bed, which includes a mattress and a synthetic pressure mat on the surface of the mattress (Fig \ref{fig:pipeline}). The human is modelled as an articulated rigid body system made with capsule primitives, which is a dynamic variant of the SMPL model. Once the simulation nears static equilibrium, we record the resting pose $\{\boldsymbol{\widetilde{\Theta}}_{C}, \tilde{\boldsymbol{\theta}}_r, \tilde{\boldsymbol{s}}_r\}$.


FleX is a position-based dynamics simulator with a unified particle representation that can efficiently simulate rigid and deformable objects. However, FleX does not currently provide a way for particles to influence the motions of rigid capsules. To overcome this limitation, we use DART \cite{lee2018dart} to model the rigid body dynamics of the capsulized human model. We combine FleX and DART through the following loop: 1) DART moves the capsulized articulated rigid body based on applied forces and moments. 2) FleX moves the soft body particles in response to the motions of the rigid body. 3) We compute new forces and moments to apply in DART based on the state of the FleX particles and the capsulized articulated rigid body. 4) Repeat. We call the combination of the two simulators DartFleX and Section \ref{ssec:dartflex} provides further details.

\textbf{Physics Simulation \#2: Pressure Image.} The settled, capsulized body is insufficient for producing a realistic pressure image: it approximates the human shape too roughly. Instead, we create a weighted, particlized, soft human body in FleX (Figs. \ref{fig:pipeline} and \ref{fig:pressurepose}) from the SMPL \cite{loper2015smpl} mesh using body shape and resting pose $\{\bm{\beta}, \boldsymbol{\widetilde{\Theta}}_{C}, \tilde{\boldsymbol{\theta}}_r \}$. 
We initialize the particlized human with 2D translation over the surface of the mattress $\{ \tilde{s}_{r,1}, \tilde{s}_{r,2} \} \in \tilde{\boldsymbol{s}}_r$. We set $s_{r,3}$, the position normal to gravity, so the body is just above the surface of the bed. We then start the simulation, resting the particlized body on the soft bed, and record the pressure image $\mathcal{P}$ once the simulation has neared static equilibrium. We note that this particlized representation has no kinematics and cannot be used to adjust a body to a resting configuration; thus our use of two separate dynamic simulations. 

\subsection{Soft Body Simulation with FleX.}\label{ssec:flex}
We simulate the sensing array by connecting FleX particles in a way that mimics real pressure sensing fabric, and model the mattress with a soft FleX object. 

\textbf{Soft Mattress and Pressure Sensing Mat.}
Here we describe the soft mattress and pressure sensing array within the FleX environment, as shown in Fig. \ref{fig:flex_dart_modeling} and further described in Appendix A.3.
The mattress is created in a common twin XL size with clusters of particles defined by their spacing, $D_M$, radius, $R_M$, stiffness, $K_M$, and particle mass, $m_M$, parameters. We then create a simulated pressure sensing mat on top of the mattress that is used to both generate pressure images and to help the human model reach a resting pose by computing the force vectors applied to the various segments of the human body. The mat consists of two layers of staggered quad FleX cloth meshes in a square pyramid structure, where each layer is defined by its stretching, $K_\sigma$, bending, $K_B$, and shear, $K_\tau$, stiffnesses, which are spring constraints on particles that hold the mat together. A compression stiffness, $K_C$, determines the bond strength between the two layers, and its mass is defined by $m_L$. 

\begin{figure}
\begin{center}

\includegraphics[width=8.2cm]{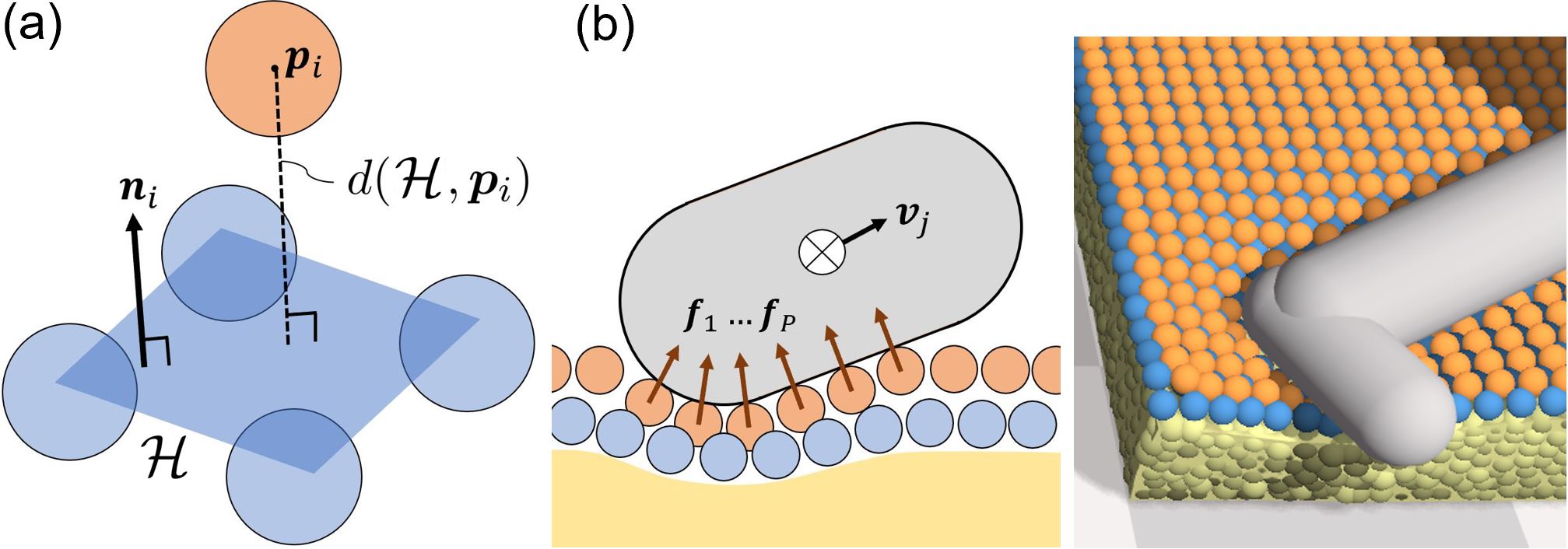}
\vspace{-4mm}
\end{center}
\caption{(a) Synthetic pressure mat structure. Pressure is a function of the penetration of the top layer array particle into the four underlying particles. (b) DartFleX collision between a capsulized limb and the simulated bed and pressure-sensing mat.}
\vspace{-3mm}
\label{fig:flex_dart_modeling}
\end{figure}

We model force applied to the mat as a function of the particle penetration vector $\boldsymbol{x}_i$ based on the pyramid structure in Fig.~\ref{fig:flex_dart_modeling} (a). Force increases as the $i^{\text{th}}$ particle on the top layer, $\boldsymbol{p}_i$, moves closer to the four particles underneath.

\begin{equation}
\boldsymbol{x}_i = \big(d_0 - d(\mathcal{H},\boldsymbol{p}_i) \big)\boldsymbol{n}_i
\label{eq:x}
\end{equation}
where $d$ is the distance between particle $\boldsymbol{p}_i$ and an approximate underlying plane $\mathcal{H}$, $d_0$ is the initial distance at rest prior to contact, and $\boldsymbol{n}_i$ is the normal vector of the approximate underlying plane.

\textbf{Sensor Model.} The BodiTrak pressure-sensing mat has an array of pressure-sensing taxels (tactile pixels). The four particles at the base of the pyramid structure in Fig.~\ref{fig:flex_dart_modeling} (a) model the 1" square geometry of a single pressure-sensing taxel. We model the pressure output, $u_i$, of a single taxel, $i$, using a quadratic function of the magnitude of the penetration vector $\boldsymbol{x}_i$.

\begin{equation}
u_i = \big(C_2|\boldsymbol{x}_i|^2 + C_1|\boldsymbol{x}_i| + C_0 \big)
\end{equation}
where $C_2$, $C_1$, and $C_0$ are constants optimized to fit calibration data, as described in Section \ref{ssec:calibration}. 

\subsection{DartFleX: Resting a Dynamic Ragdoll Body}\label{ssec:dartflex}

The purpose of DartFleX is to allow rigid body kinematic chains to interact with soft objects by coupling the rigid body dynamics solver in DART to the unified particle solver in FleX as shown in Fig. \ref{fig:flex_dart_modeling} (b). 

\textbf{Dynamic rigid body chain.} Our rigid human body model relies on a capsulized approximation to the SMPL model, following \cite{bogo2016keep}. To use this model in a dynamics context, we calculate the per-capsule mass 
based on volume ratios from a person with average body shape $\bar{\boldsymbol{\beta}} = \mathbf{0}$, average body mass, and mass percentage distributions between body parts as defined by Tozeren \cite{tozeren1999human}. For joint stiffnesses $\boldsymbol{k}_{\theta} \in \mathbbm{R}^{69}$, we tune parameters to achieve the low stiffness characteristics of a ragdoll model that can settle into a resting pose on a bed due to gravity. We set torso and head stiffness high so that they are effectively immobile, and joint damping $\boldsymbol{b}_{\theta} = 15\boldsymbol{k}_{\theta}$ to reduce jitter.

\textbf{DartFleX Physics.} 
We initialize the same capsulized model in both DART and FleX. We apply gravity in DART, and take a step in the DART simulator. We get a set of updated dynamic capsule positions and orientations, and move the static geometry counterparts in FleX accordingly. In order to transfer force data from FleX to DART, we first check if any top layer pressure mat particles are in contact. Each particle $i$ in contact has a penetration vector $\boldsymbol{x}_i(t)$ (see equation \ref{eq:x}) at time $t$, which we convert to normal force vector $\boldsymbol{f}_{N,i} \in \mathbbm{R}^{3}$ using a mass-spring-damper model \cite{payandeh2001finite}: 

\begin{equation}
\boldsymbol{f}_{N,i} = k \boldsymbol{x}_i(t) + b \boldsymbol{\dot{x}}_i(t),
\label{eq:msd}
\end{equation}
where $k$ is a spring constant, $b$ is a damping constant, and $\boldsymbol{f}_{N,i} \perp \mathcal{H}$. 
We then assign each force to its nearest corresponding capsule $j$. Given the velocity, $\boldsymbol{v}_j$, of capsule $j$ and a friction coefficient, $\mu_k$, we compute the frictional force $\boldsymbol{f}_{T,i}$ for the $i^{th}$ particle in contact:

\begin{equation}
\boldsymbol{f}_{T, i} = -\mu_k|\boldsymbol{f}_{N,i}|{\boldsymbol{v}_j - proj_{\boldsymbol{f}^{}_{N, i}}\boldsymbol{v}_j  \over |\boldsymbol{v}_j - proj_{\boldsymbol{f}^{}_{N, i}}\boldsymbol{v}_j |}
\end{equation}
where $proj$ is an operator that projects $\boldsymbol{v}_j$ orthogonally onto a straight line parallel to $\boldsymbol{f}^{}_{N, i}$ . 
In our simulation, we set $b = 4k$ and $u_k = 0.5$, and we find $k$ through a calibration sequence described in Section \ref{ssec:calibration}. We can then compute the total particle force, $\boldsymbol{f}_{i}$:

\begin{equation}
\boldsymbol{f}_{i} = \boldsymbol{f}_{N,i} + \boldsymbol{f}_{T,i}
\end{equation}

We then compute a resultant force $\boldsymbol{F}_{j}$ in FleX for the $j^{th}$ body capsule, based on the sum of forces from $P$ particles in contact with the capsule plus gravity, $\boldsymbol{F}_{g}$:

\begin{equation}
\boldsymbol{F}_{j} = \sum_{i=1}^P{\boldsymbol{f}_i} + \boldsymbol{F}_{g}
\end{equation}
Moment $\boldsymbol{M}_{j}$ is computed on each capsule $j$ from $P$ particles in contact, where $\boldsymbol{r}_i$ is the moment arm between a particle and the capsule center of mass:  

\begin{figure}
\begin{center}
\includegraphics[width=8.2cm]{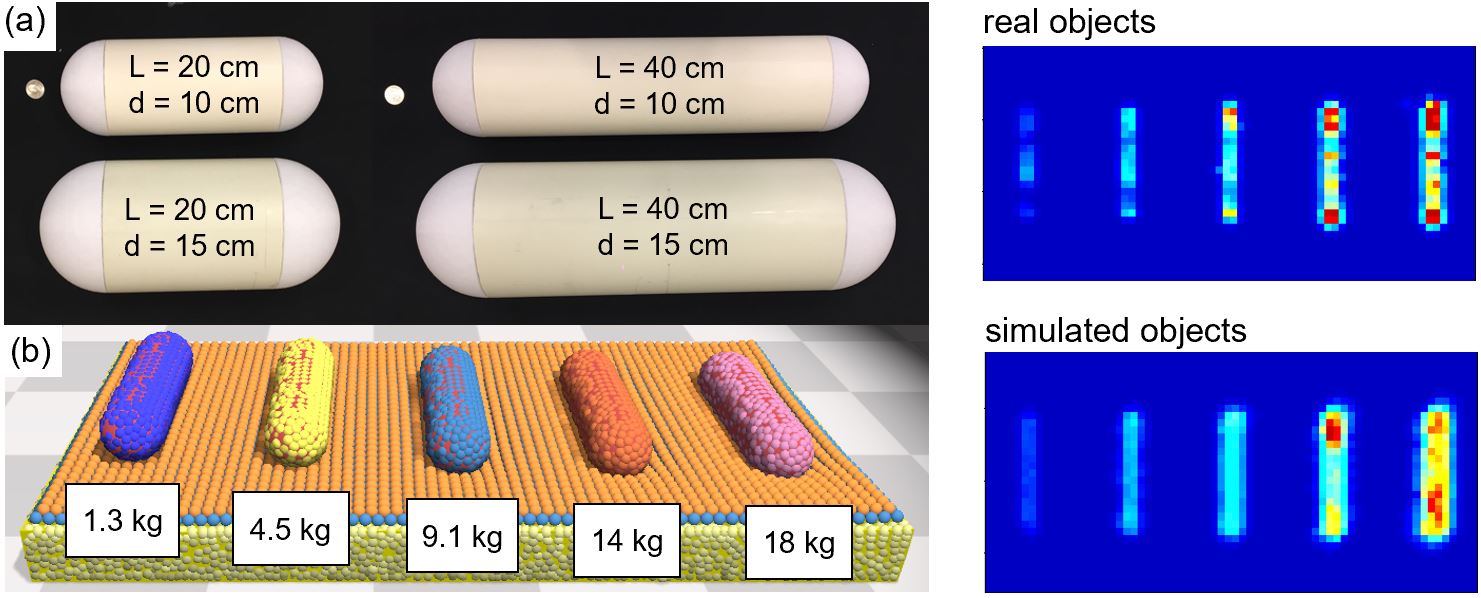}
\vspace{-3mm}
\end{center}
\caption{(a) Rigid calibration capsules with quarters (U.S. coins) shown for size. (b) Simulated capsules. (right) Real and simulated pressure images prior to calibration.}
\vspace{-3mm}
\label{fig:calibration_shapes}
\end{figure}

\begin{equation}
\boldsymbol{M}_{j} = \sum_{i=1}^P{\boldsymbol{r}_{i} \times\boldsymbol{f}_i}
\end{equation}
The resultant forces and moments are applied in DART, a step is taken with the forces and gravity applied to each body part, and the DartFleX cycle repeats. We continue until the capsulized model settles and then record resting pose $\boldsymbol{\widetilde{\Theta}}_{C}$, root position $\tilde{\boldsymbol{s}}_r$, and root orientation $\tilde{\boldsymbol{\theta}}_r$.

\begin{figure*}
\begin{center}
\includegraphics[width=\textwidth]{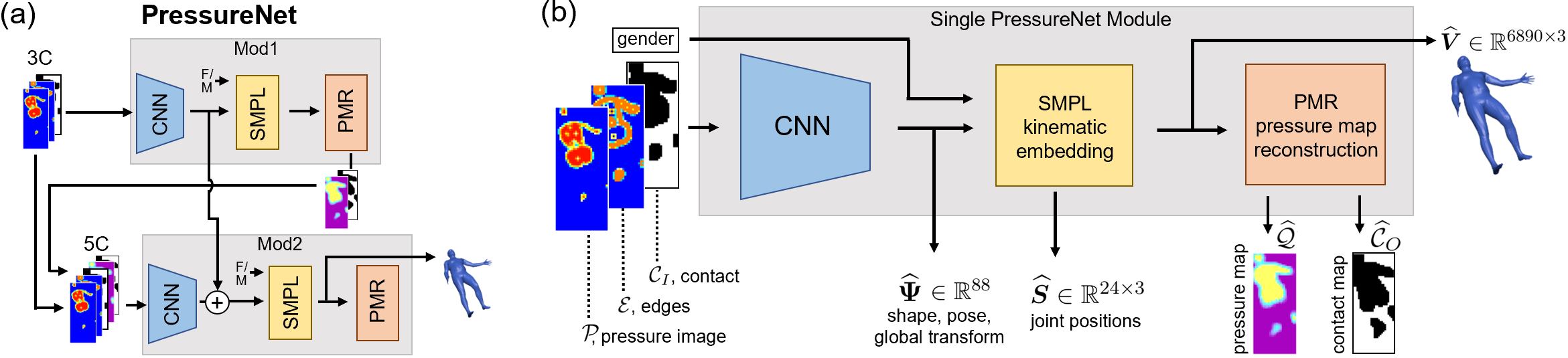}
\end{center}
\vspace{-4mm}
\caption{\label{fig:network_wide}(a) PressureNet: We combine two network modules (``Mod1'' and ``Mod2'') in series. Mod1 learns a coarse estimate and Mod2 fine-tunes, by learning a residual that takes as input the two maps reconstructed by Mod1 combined with the input to Mod1.  (b) Detailed description of a single PressureNet module showing the novel PMR network that reconstructs pressure and contact maps.}
\vspace{-3mm}
\end{figure*}

\subsection{Calibration}\label{ssec:calibration}
We calibrated our simulation using the rigid capsule shapes in Fig.~\ref{fig:calibration_shapes} (a). We placed varying weights on them on the real pressure-sensing mat and recorded data, and then created matching shapes in simulation. We first calibrated the FleX environment using the particlized capsules shown in Fig.~\ref{fig:calibration_shapes} (b) using the covariance matrix adaptation evolution strategy (CMA-ES) \cite{hansen2019pycma} to match synthetic pressure images and real pressures images of the calibrated objects by optimizing $D_M$, $R_M$, $K_M$, $m_M$, $d_0$, $K_{\sigma}$, $K_B$, $K_{\tau}$, $K_C$, $m_L$, $C_2$, $C_1$, and $C_0$. 

We also measure how much the real capsules sink within the mattress. We use these measurements to calibrate the mass-spring-damper model in equation \ref{eq:msd}. We fit the simulated capsule displacement to the real capsule displacement to solve for the spring constant $k$ and then set $b = 4k$ and $\mu_k = 0.5$. See Appendix A.4 and A.5 for details. 

\section{PressureNet}\label{sec:methods2} Given a pressure image of a person resting in bed and a gender, PressureNet produces a posed 3D body model.
PressureNet (Fig.~\ref{fig:network_wide}~(a)) consists of two network modules trained in sequence (``Mod1'' and ``Mod2''). Each takes as input a tensor consisting of three channels: pressure, edges, and contact $\{\mathcal{P}, \mathcal{E}, \mathcal{C}_I\}\in\mathbb{R}^{128 \times 54 \times 3}$, which are shown in Fig.~\ref{fig:network_wide}~(b), as well as a binary flag for gender. $\mathcal{P}$ is the pressure image from a pressure sensing mat, $\mathcal{E}$ results from an edge detection channel consisting of a sobel filter applied to $\mathcal{P}$, and $\mathcal{C}_I$ is a binary contact map calculated from all non-zero elements of $\mathcal{P}$. Given this input, each module outputs both an SMPL mesh body and two reconstructed maps produced by the PMR network, $\{ \widehat{\mathcal{Q}}, \widehat{\mathcal{C}}_{O} \}$, that estimate the pressure image that would be generated by the mesh body. Mod2 has the same structure as Mod1, except that it takes in two additional channels: the maps produced by PMR in Mod1 $\{ \widehat{\mathcal{Q}}_1, \widehat{\mathcal{C}}_{O,1} \}$. We train PressureNet by training Mod1 to produce a coarse estimate, freezing the learned model weights, and then training Mod2 to fine-tune the estimate. 

\textbf{CNN.} The first component of each network module is a CNN with an architecture similar to the one proposed by Clever et al~\cite{clever20183d}. Notably, we tripled the number of channels in each convolutional layer. See Appendix B.1 for details. During training, only the weights of the CNNs are allowed to change. All other parts of the networks are held constant. The convolutional model outputs the estimated body shape, pose, and global transform, $\bm{\widehat{\Psi}} = \{\bm{\widehat{\Theta}}, \bm{\hat{\beta}}, \bm{\hat{s}}_r, \bm{\hat{x}}_r, \bm{\hat{y}}_r\}$, with the estimated joint angles $\bm{\widehat{\Theta}} \in \mathbbm{R}^{69}$, body shape parameters $\bm{\hat{\beta}} \in \mathbbm{R}^{10}$, global translation of the root joint with respect to the bed $\bm{\hat{s}}_r \in \mathbbm{R}^{3}$, and parameters $\bm{\hat{x}}_r, \bm{\hat{y}}_r$ which define a continuous orientation for the root joint of the body with $\{x_{u}, x_{v}, x_{w}\} \in \boldsymbol{x}_r$,  $\{y_{u}, y_{v}, y_{w}\} \in \boldsymbol{y}_r$ for 3 DOF,  i.e. $\theta_{r,u}  = atan2(y_{u}, x_{u})$ and $\{\theta_{r,u}, \theta_{r,v}, \theta_{r,w} \} \in \boldsymbol{\theta}_r \in \mathbbm{R}^{3}$. 

\textbf{SMPL kinematic embedding.} $\bm{\widehat{\Psi}}$ feeds into a kinematic embedding layer (see Fig.~\ref{fig:network_wide}), which uses the SMPL differentiable kinematics model from \cite{kanazawa2018end} to learn to estimate the shape, pose, and global transform. This embedding outputs joint positions for the human body, $\bm{\widehat{S}}$, and a SMPL mesh consisting of vertices $\boldsymbol{\widehat{V}}$; and relies on forward kinematics to ensure body proportions and joint angles match real humans. 

\textbf{PMR.} The final component of each module, the PMR network, reconstructs two maps based on the relationship between the SMPL mesh $\boldsymbol{\widehat{V}}$ and the surface of the bed. The reconstructed pressure map ($\widehat{\mathcal{Q}}$) corresponds with the input pressure image, $\mathcal{P}$, and is computed for each pressure image taxel based on the distance that the human mesh sinks into the bed. The reconstructed contact map ($\widehat{\mathcal{C}}_{O}$) corresponds with the input contact map, $\widehat{\mathcal{C}}_{I}$, and is a binary contact map of $\widehat{\mathcal{Q}}$. See Appendix B for details.

\textbf{Loss function.} We train Mod1 in PressureNet with the following loss function, given $N=24$ Cartesian joint positions and $S=10$ body parameters: 

\setlength{\abovedisplayskip}{-10pt}
\setlength{\belowdisplayskip}{5pt}

\begin{equation}
\mathbbm{L}_1 = 
 {1\over N\sigma_s}\sum_{j=1}^{N}\big|\big|\bm{s}_j - \bm{\hat{s}}_{j,1}\big|\big|^{}_2
+ {1\over  S\sigma_\beta}\big|\big|\bm{\beta} - \bm{\hat{\beta}}_1\big|\big|^{}_1
\end{equation}
where $\bm{s}_j \in \bm{{S}}$ represents the 3D position of a single joint, and $\sigma_{s}$ and $\sigma_{\beta}$ are standard deviations computed over the whole dataset to normalize the terms. 

In our evaluations (Section~\ref{sec:results}), sequentially training two separate network modules improved model performance and the resulting human mesh and pose predictions.
For a pressure array of $T$ taxels, we compute a loss for Mod2 by adding the error between the reconstructed pressure maps and the ground truth maps from simulation.

\begin{equation}
\mathbbm{L}_{2} = \mathbbm{L}_{1} + {1\over T\sigma^{}_{\mathcal{Q}}}\big|\big|\mathcal{Q} - \widehat{\mathcal{Q}}_2\big|\big|^2_2
+ {1\over T\sigma^{}_{\mathcal{C}_O}}\big|\big|\mathcal{C}_O - \widehat{\mathcal{C}}_{O,2}\big|\big|^{}_1
\end{equation}
where $\mathbb{L}_1$ uses Mod2 estimates (i.e. $\bm{\widehat{S}}_2$, $\bm{\hat{\beta}}_2$), $\mathcal{Q}$ and $\mathcal{C}_O$ are ground truth maps precomputed by setting $\bm{\widehat{\Psi}} = \bm{\Psi}$, and $\sigma^{}_{\mathcal{Q}}$, $\sigma^{}_{\mathcal{C}_O}$ are computed over the dataset.

\section{Evaluation}\label{sec:evaluation}
To evaluate our methods, we trained our CNN on synthetic data and tested it on both synthetic and real data. We generated 206K synthetic bodies at rest with corresponding pressure images (184K train / 22K test), which we partitioned to represent both a uniformly sampled space and common resting postures. By posture, we mean common recognized categories of overall body pose, such as sitting, prone, and supine. We tested 4 network types and 2 training data sets of different size.


\subsection{PressurePose Data Partitions}\label{ssec:eval_p1}
We used the rejection sampling method described in Section~\ref{sec:methods1} and Appendix A.1 to generate initial poses and create dataset partitions. Our main partition, the \textit{general} partition, consists of 116K image and label pairs. In it, we evenly distributed limb poses about the Cartesian space and randomly sampled over body roll and yaw. This partition includes supine, left/right lateral and prone postures, as well as postures in between, and has the greatest diversity of poses. We also created a \textit{general supine} partition (58K) featuring only supine postures and evenly distributed limb poses. Finally, we generated smaller partitions representing other common postures: \textit{hands behind the head} (5K), \textit{prone with hands up} (9K), \textit{supine crossed legs} (9K), and \textit{supine straight limbs} (9K). See Appendix A.7 for details.


\subsection{PressureNet Evaluation}\label{ssec:pnet_eval}

We normalized all input data by a per-image sum of taxels. We blurred synthetic and real images with a Gaussian of $\sigma = 0.5$. We trained for 100 epochs on Mod1 with loss function $\mathbbm{L}_1$. Then, we pre-computed the reconstruction maps $\{ \widehat{\mathcal{Q}}_1, \widehat{\mathcal{C}}_{O,1} \}$ from Mod1 for input to Mod2, and trained Mod2 for 100 epochs using loss function $\mathbbm{L}_2$. See Appendix B.3 for training hyperparameters and details. 

We investigated 5 variants of PressureNet, which are all trained entirely with synthetic data in order to compare the effect of (1) ablating PMR, (2) adding noise to the synthetic training data, (3) ablating the contact and edge input ( $\mathcal{C}_I$ and  $\mathcal{E}$ ), and (4) reducing the training data size. Ablating PMR consists of removing the 2 reconstructed maps from the input to Mod2 and using $\mathbb{L}_1$ for training both Mod1 and Mod2. We compared the effect of adding noise to the training data to account for real-world variation, such as sensor noise. Our noise model includes per-pixel white noise, additive noise, multiplicative noise, and blur variation, all with $\sigma = 0.2$. We compared networks trained on 46K vs. 184K images. 



\subsection{Human Participant Study}
We mounted a Microsoft Kinect 2 $1.6 m$ above our Invacare Homecare bed to capture RGB images and point clouds synchronized with our pressure image data. See details in Appendix A.6.
We recruited 20 (10F/10M) human participants with approval from an Institutional Review Board. We conducted the study in two parts to capture (1) participant-selected poses and (2) prescribed poses from the synthetic test set. We began by capturing five participant-selected poses. For the first pose, participants were instructed to get into the bed and get comfortable. For the remaining four, participants were told to get comfortable in supine, right lateral, left lateral, and prone postures. Next, for the prescribed poses, we displayed a pose rendering on a monitor, and instructed the participants to get into the pose shown.  We captured 48 prescribed poses per participant, which were sampled without replacement from the synthetic testing set: 24 general partition poses, 8 supine-only poses, and 4 from each of the remaining partitions. 

\begin{figure}
\begin{center}
\includegraphics[width=8.2cm]{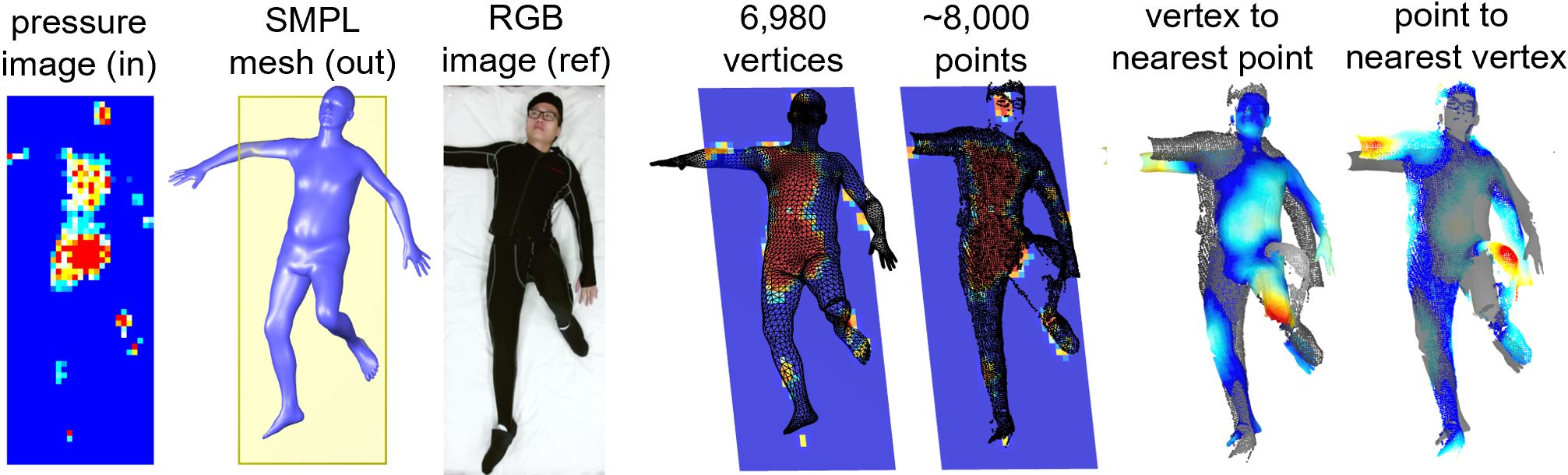}
\vspace{-4mm}
\end{center}
    \caption{3D error analysis between a human mesh (6,980 vertices) and a point cloud ($\sim$8,000 downsampled points).}
 \vspace{-3mm}
\label{fig:error_analysis}
\end{figure}

\subsection{Data Analysis}

We performed an error analysis as depicted in Fig. \ref{fig:error_analysis}. For this analysis, we compute the closest point cloud point to each mesh vertex, and the closest mesh vertex to each point cloud point. We introduce 3DVPE (3D vertex-point-error), which is the average of these numbers. We downsample the point cloud to a resolution of 1cm so the number of points is roughly equal to the number of mesh vertices. We clip the mesh vertices and the point cloud at the edges of the pressure mat. The point cloud only contains information from the top surface of the body facing the camera, so we clip the mesh vertices that do not have at least one adjacent face facing the camera. Finally, we normalize the mesh by vertex density: while the density of the point cloud is uniform from downsampling, the mesh vertices are highly concentrated in some areas like the face. We normalize each per-vertex error by the average of its adjacent face surface areas.

\begin{figure*}
\begin{center}
\includegraphics[width=17.5cm]{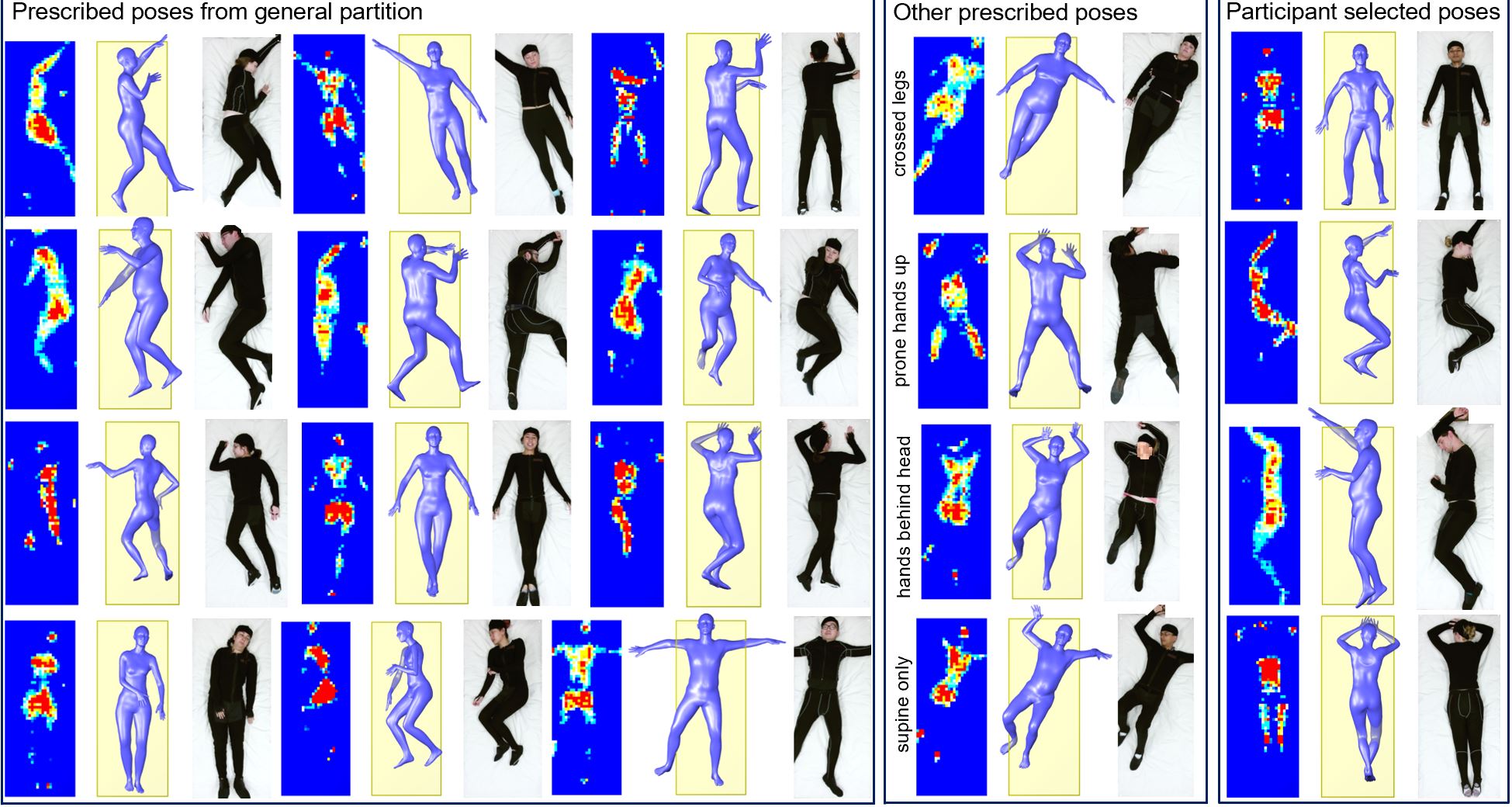}
\end{center}
\vspace{-5mm}
    \caption{PressureNet results on real data with the best performing network (trained with 184K samples).}
\vspace{-4mm}
\label{fig:big_results}
\end{figure*}

We evaluated PressureNet on the synthetic test set and compared the results to the real test set. We clip the estimated and ground truth mesh vertices and normalize per-vertex error in the same way as the real data. Additionally, we evaluated per-joint error ($24$ joints) using mean-per-joint-position error (MPJPE), and per-vertex error ($6890$ vertices) using vertex-to-vertex error (v2v) for the synthetic data. We evaluated the network's ability to infer posture using the participant-selected pose dataset by manually labeling the inferred posture (4 labels: supine, prone, left/right lateral). We also compared to a baseline human, \textit{BL}, where we put a body of mean shape in a supine position in the center of the bed and compare it to all ground truth poses. We positioned the legs and arms to be straight and aligned with the length of the body.

\section{Results and Discussion}\label{sec:results}

\begin{table}[t!]
\begin{center}

\renewcommand{\arraystretch}{1.1}

\scalebox{0.7}{\begin{tabular} {c|c|c|c|c|c|c}
\hline
 Network Description&   
 \parbox[t]{3mm}{{\rotatebox[origin=c]{90}{Training}}}
 \parbox[t]{2mm}{{\rotatebox[origin=c]{90}{data ct.}}}&  
 \parbox[t]{3mm}{{\rotatebox[origin=c]{90}{12K synth}}}
 \parbox[t]{2mm}{{\rotatebox[origin=c]{90}{ MPJPE (cm) }}}& 
 \parbox[t]{3mm}{{\rotatebox[origin=c]{90}{12K synth}}}
 \parbox[t]{2mm}{{\rotatebox[origin=c]{90}{v2v (cm)}}}& 
 \parbox[t]{3mm}{{\rotatebox[origin=c]{90}{12K synth}}}
 \parbox[t]{2mm}{{\rotatebox[origin=c]{90}{ 3DVPE (cm) }}}&  
 \parbox[t]{3mm}{{\rotatebox[origin=c]{90}{1K real}}}
 \parbox[t]{2mm}{{\rotatebox[origin=c]{90}{ 3DVPE (cm) }}}& 
 \parbox[t]{3mm}{{\rotatebox[origin=c]{90}{99 real}}}
 \parbox[t]{2mm}{{\rotatebox[origin=c]{90}{ 3DVPE (cm) }}}\\

\hline
\hline
Best & 184K &\textbf{11.18} & \textbf{13.50} & \textbf{3.94} & \textbf{4.99} & \textbf{4.76} \\
Noise $\sigma$ ablated & 184K & \textbf{11.18} & 13.52 & 3.97 & 5.05 & 4.79 \\
 Input $\mathcal{E}$, $\mathcal{C}_I$ ablated & 184K &11.39 & 13.73 & 4.03 & 5.07 & 4.85 \\
 Best - small data & 46K & 12.65 & 15.28 & 4.35 & 5.17 & 4.89 \\
 PMR ablated & 184K & 12.28 & 14.65 & 4.38 & 5.33 & 4.94 \\
Baseline - mean pose & - & 33.30 & 38.70 & 8.43 & 6.65 & 5.22\\
\hline
\cline{1-7}

\end{tabular}}
\end{center}
\vspace{-0.2cm}
\caption{\label{tab:results_table}Results comparing testing data and network type. 
}
\vspace{-0.2cm}
\end{table}

Overall, we found that using more synthetic data resulted in higher performance in all tests, as shown in Table \ref{tab:results_table}. As expected, ablating the PMR network and ablating noise reduced performance. Ablating contact and edge inputs also reduced performance. We expect that comparable performance could be achieved without them, possibly by changing the details of the CNN. Fig.~\ref{fig:big_results} shows results from the best performing network with 184K training images, noise, and the PMR network.



We compared the error on a set of 99 participant selected poses, shown in Table \ref{tab:results_partselect}, using the best performing PressureNet. Results show a higher error for lateral postures where the body center of mass is further from the mat and limbs are more often resting on other limbs or the body rather than the mat. Results on partitioned subsets of data can be found in Appendix B.4. Fig.~\ref{fig:failure_cases_main} shows four failure cases.


\begin{table}[t!]
\begin{center}
\vspace{-0.15cm}
\renewcommand{\arraystretch}{1.1}
\vspace{1mm}

\scalebox{0.75}{\begin{tabular} {c|c|c|c}
\hline
posture partition & test ct. & 3DVPE (cm) & posture match \\ 
\hline
\hline

no instruction & 19 & 3.93 & 100\% \\
supine & 20 & 4.02  & 100\% \\
right lateral & 20 & 5.45  & 100\% \\
left lateral & 20 & 5.37 & 100\% \\
prone & 20 & 4.96  & 95\%* \\
\hline
\cline{1-4}
\end{tabular}}
\end{center}
\vspace{-0.2cm}
\caption{\label{tab:results_partselect}Results - participant selected poses. *See Fig.~\ref{fig:failure_cases_main}-top left.}
\vspace{-0.2cm}
\end{table}

\begin{figure}
\begin{center}
\includegraphics[width=7.6cm]{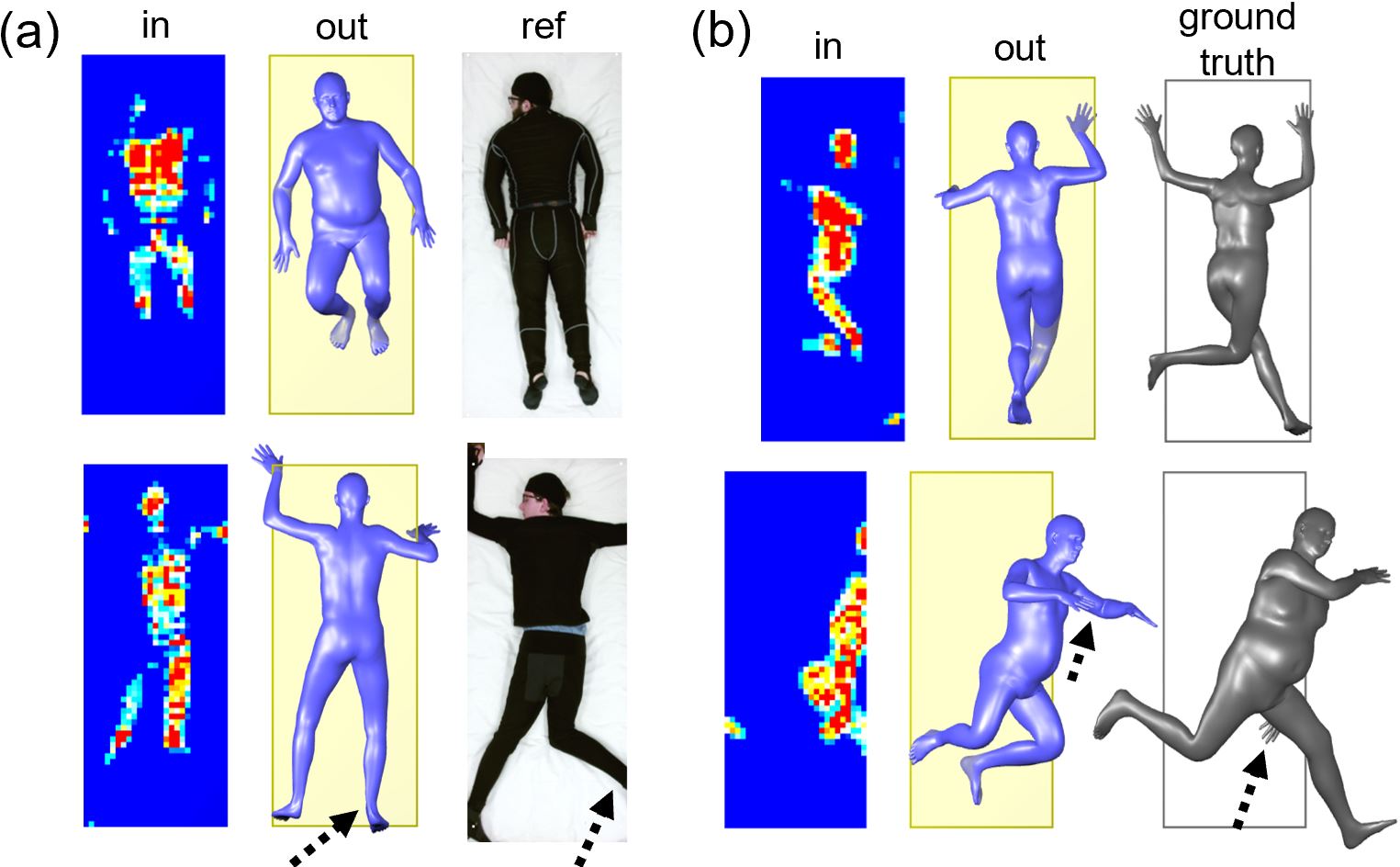}
\vspace{-4mm}
\end{center}
    \caption{Some failure cases. (a) Real data. (b) Testing on synthetic \textit{training} data.}
 \vspace{-5mm}
\label{fig:failure_cases_main}
\end{figure}


\section{Conclusion}

With our physics-based simulation pipeline, we generated a dataset, PressurePose, consisting of 200K synthetic pressure images with an unprecedented variety of body shapes and poses. Then, we trained a deep learning model, PressureNet, entirely on synthetic data. With our best performing model, we achieve an average pose estimation error of $< 5~cm$, as measured by 3DVPE, resulting in accurate 3D pose and body shape estimation with real people on a pressure sensing bed.

\textbf{Acknowledgement:} We thank Alex Clegg. This work was supported by the National Science Foundation Graduate Research Fellowship Program under Grant No. DGE-1148903, NSF award IIS-1514258, NSF award DGE-1545287 and AWS Cloud Credits for Research.

\textbf{Disclosure:} Charles C. Kemp owns equity in and works for Hello Robot, a company commercializing robotic assistance technologies. Henry M. Clever is entitled to royalties derived from Hello Robot's sale of products.

\begin{alphasection}
\section*{Appendix A: PressurePose Data Generation}

\subsection{Initial Pose Sampling}\label{ssec:init_sampling}
We use rejection sampling to generate initial pose dataset partitions. Our criteria are as follows.

\textbf{Uniform Cartesian space distribution} - Fig. \ref{fig:rejection_sampling} (a). We use rejection sampling to uniformly sample poses with respect to the Cartesian space, by discretizing the space and ensuring that a given limb is equally represented in each unit. We define a Cartesian space $\bm{\mathcal{Y}}$ as a cuboid for checking for presence of the most distal limb. First, we constrain $\bm{\mathcal{Y}}$ in the $(x,y)$ directions to how far the distal joint (e.g. right foot, $\bm{s}_{r.foot}$) can extend from the promixal joint (e.g. right hip, $\bm{s}_{r.hip}$) in a limb. For the legs, we assume that the foot cannot move above the hip. For the right leg, these constraints can be summarized as: $s_{r.foot,x} \in [s_{r.hip, x} - l_{leg}, s_{r.hip, x} + l_{leg}]$ and $s_{r.foot,y} \in [s_{r.hip, y}, s_{r.hip, y} + l_{leg}]$. We also constrain the $z$ direction to ensure that the distal joint is initially positioned at a height close to where the proximal joint is: For laying poses, the distal joints (feet and hands) are more likely to end up close to the surface of the bed than very high in the air, for example. This constraint promotes simulation stability and decreases the time it takes for physics simulation \#1 (Fig. 2) to reach an equilibrium state. We constrain $s_{r.foot,z} \in [s_{r.hip, z} - 10cm, s_{r.hip, z} + 10cm]$. 

Next, we break up $\bm{\mathcal{Y}}$ into a set of smaller cuboids as shown in Fig.~\ref{fig:rejection_sampling}-top middle. For each limb we uniformly sample a cuboid from $\{\mathcal{Y}_1, ... \}$ and then use rejection sampling on the limb joint angles --- in the case of Fig. \ref{fig:rejection_sampling} (a), the right leg --- until $\bm{s}_{r.foot} \in \mathcal{Y}_4$.

\textbf{Generate common posture partitions} - Fig.~\ref{fig:rejection_sampling} (b). Some common postures, such as resting with the hands behind the head, are unlikely to be generated when the joint angles are sampled from a uniform distribution. For example, there is a $<1\%$ probability of generating a pose with the hands-behind-the-head when sampling joint angles uniformly, so a network trained with such little hands-behind-the-head data has difficulty learning such a pose. We mitigate this issue by checking for presence of the most distal joint in a cuboid representing where it would be located in such as pose. If the joint is within the cuboid, e.g. $\bm{s}_{r.hand} \in \mathcal{Y}_{RH}$, the joint passes the criteria and we add the limb pose to the set of checked initial poses.

\begin{figure}[t]
\begin{center}
   \includegraphics[width=1.0\linewidth]{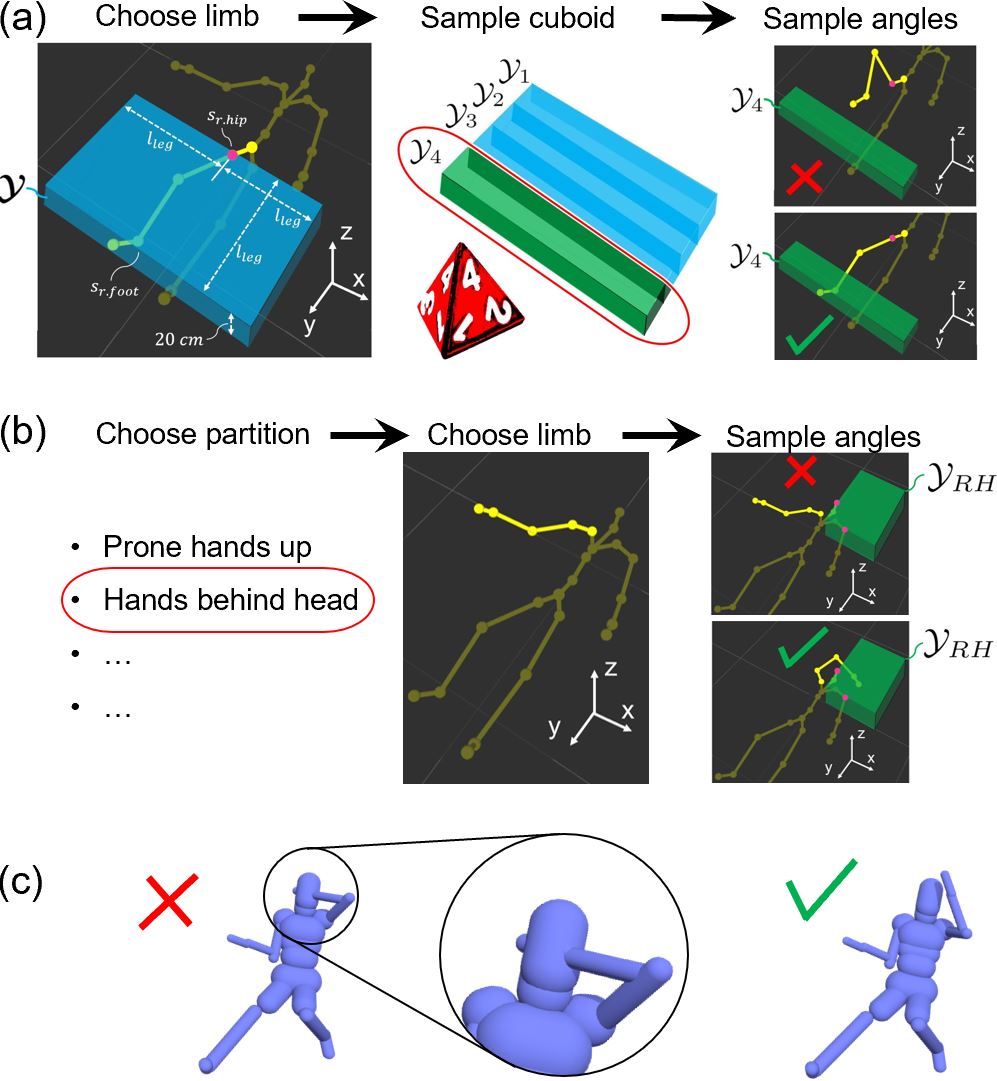}
\end{center}
\vspace{-0.3cm}
   \caption{Rejection sampling criteria. (a) Evenly distributing right leg poses across Cartesian space by sampling from four non-overlapping Cartesian cuboids, $\{\mathcal{Y}_1, \mathcal{Y}_2, \mathcal{Y}_3, \mathcal{Y}_4, \} \in \bm{\mathcal{Y}}$. Reject pose angles if $\boldsymbol{s}_{r.ankle}\not\in \mathcal{Y}_4$  (b) For sampling right arm in the hands-behind-head partition, we reject the right arm pose angles if $\boldsymbol{s}_{r.hand}\not\in \mathcal{Y}_{RH}$. (c) Pose feasibility checking via collision detection.}
\label{fig:rejection_sampling}
\vspace{-0.5cm}
\end{figure}

\textbf{Prevent self-collision} - Fig. \ref{fig:rejection_sampling} (c). We reject poses that result in self collision by capsulizing the mesh and using the DART collision detector. We check the hands, forearms, feet, and lower leg capsules for collision with any other capsules except their adjacent capsules (e.g. forearm and upper arm should overlap).

\subsection{Dynamic Simulation Details}
\textbf{Weighting particles in FleX}. We directly calculate particle mass for the particlized human in physics simulation \#2, as well as for the particlized calibration objects depicted in Fig. 5 (b). Since FleX is a position-based dynamics simulator and the mass is defined by units of inverse mass $1 / m$ on an arbitrary scale, we begin by defining the inverse mass scale for particles in the particlized human. 

For this, we assume that the volume each particle in the human takes up, as well as the density of particles, is the same for that of water. 
Because volume and density are equal, we also can set inverse mass equal, so $1 / m_{H} = 1$, thus $1 / m_{H2O} = 1$.

We calculate the inverse mass for particles in calibration objects by a density ratio to that of water, given a known weight of the object $w_k$ and the object volume $V_o$: 

\begin{equation}
{1 \over m_{o, k}} = {1 \over m^{}_{H2O}}{\rho^{}_{H2O} \over \rho_{o}} = {\rho^{}_{H2O} \over m^{}_{H2O}}{V_o \over w_k /g}
\end{equation}
where $\rho^{}_{H2O}$ is the density of water and $g$ is gravity. In contrast to the humans and objects rested on the bed, the the soft mattress and synthetic pressure mat particle inverse mass are determined from an optimization described in Appendix \ref{ssec:flexcal}.

\textbf{Weighting the capsulized human chain.} We compute a per-capsule weight for the articulated capsulized chain in DartFleX based on the weight distribution for an average person and capsule volume ratios. First, we describe how we assign capsule mass for the average person. We use average body mass and mass distribution values from Tozeren \cite{tozeren1999human}, and calculate capsule volumes from body shape. We assume the average human of gender $g\in \{M, F\}$ has a mass of $\bar{m_g}$, mass percentage distribution for body part $R$ of $\bar{X}_{R,g} \in \bar{\boldsymbol{X}}_{g}$, and SMPL body shape parameters $\bar{\boldsymbol{\beta}_g} = \mathbf{0}$. We define the mass of each capsule $c$ in an average person to be:

\begin{equation}
{\bar{m_c} = \bar{m_g} \bar{X}_{R,g} {\bar{V}_{c,g} \over \bar{V}_{R,g}}}
\end{equation}
where $\bar{V}_{c,g}$ is the volume of capsule $c$ for a mean body shape $\bar{\boldsymbol{\beta}_g}$, and $ \bar{V}_{R,g}$ is the sum of volumes for all capsules in body part $R$. Now, we describe how this capsule mass can be converted into masses for people of other shapes. To find the mass of some capsule $c$ for a body of particular shape $\boldsymbol{\beta}$, we use a capsule volume ratio between the particular person and an average person:

\begin{equation}{
m_c = \bar{m_c} {V_c \over \bar{V}_{c,g}}}
\end{equation}
where $V_c$ is the volume of some arbitrary capsule. Computing capsule volume analytically is simple given radius and length, but this is complicated by capsule overlap, which is often substantial in the SMPLIFY capsulized model \cite{bogo2016keep} we use. Instead, we use discretization to compute capsule volume and correct for overlap. First, we use the SMPLIFY regressor to calculate capsule radius and length from body shape $\boldsymbol{\beta}$. Besides shape, overlap is dependent on the particular pose of the capsulized model. We assume that pose dependent differences in overlap are very small, and set the pose constant at $\boldsymbol{\Theta} = \boldsymbol{0}$. We then compute the global transform for each capsule using this shape and pose. From capsule radii, lengths, and global transforms, we place all capsules in 3D space and voxelize them with a resolution of $2 mm$. This produces a set of 3D masks, which are tagged to their corresponding capsules. Voxels belonging to a unique capsule are allocated directly, while voxels belonging to multiple capsules are allocated fractionally based on the number of capsules sharing the voxel. We compute capsule mass inertia matrices analytically from capsule radius and length.

\textbf{Capsulized body joint stiffness}. For an average person, we set the following joint stiffnesses for the shoulders, elbows, hands, hips, knees and feet to low stiffness: $\bar{k}_{\theta,shd} = 4$ Nm, $\bar{k}_{\theta,elb} = 2$ Nm, $\bar{k}_{\theta,hnd} = 4$ Nm, $\bar{k}_{\theta,hip} = 6$ Nm, $\bar{k}_{\theta,knee} = 3 $ Nm and $\bar{k}_{\theta,feet} = 6$ Nm. We set torso and head stiffness very stiff $\bar{k}_{\theta, trs}, \bar{k}_{\theta, hd} = 200$ Nm. For a person of particular body shape, we weight joint stiffnesses $\boldsymbol{k}_{\theta}$ by the body mass ratio, where $\boldsymbol{k}_{\theta} = (m / \bar{m}) \boldsymbol{\bar{k}}_{\theta} $. We set joint damping $\boldsymbol{b}_{\theta} = 15\boldsymbol{k}_{\theta}$. The direction and magnitude of stiffness force on each joint is dependent on joint equilibrium position, i.e. the joint angle where force is 0. We set the equilibrium position of the joints to be the \textit{home pose}, where the arms are at the sides and the legs are straight. In the SMPLIFY model, \textit{home pose} consists of equilibrium joint positions $\boldsymbol{\Theta}_{eq}$ set to 0, except the shoulders, which are bent downward at $90$ degrees. Rather than set $\boldsymbol{\Theta}_{eq}$ to initial joint angles $\boldsymbol{\Theta}_C$, we do this to guide the pose away from extreme angles at a modest force.

Because we set the joint stiffness low, our dataset does not capture non-resting postures, such when a person is getting in/out of bed (recall Table 1). However, we have been able to generate resting sitting poses by bending the mattress and pressure mat into a sitting configuration and then resting a person on it, like the sitting postures in~\cite{clever20183d}.

\textbf{Settling criteria - Physics simulation \#1}. For physics simulation \#1, the goal is to slowly allow the body to fall on the bed and settle into a resting pose. We start the capsulized body at a height based on the lowest point on the body. For many randomly sampled poses, the lowest joint is initially much lower than the center of mass, which causes the center of mass to build significant momentum by the time it reaches the bed. We found that this caused bouncing and instability, and was qualitatively different from the motion one might take to assume a resting pose in bed. We alleviate this issue by zeroing the velocity of the capsulized model every 4 iterations in the simulation ($\sim 0.04 s$) until a capsule that better represents the center of mass contacts the surface of the bed. For this, we use the capsule approximating the buttocks.

Finding a resting pose in static equilibrium is hampered by the stability of DartFleX: DART uses a more traditional physics solver and FleX uses position-based dynamics, which are challenging to connect in a stable loop. Rather than run the simulation until static equilibrium, we use a cutoff threshold that takes velocity and acceleration of all capsules into account. We define a resting body as that when the maximum velocity of all capsules has reached $v_{max} < 0.05 m /s$ and maximum acceleration has reached $a_{max} < 0.5 m / s^2$. In the event the model does not settle within $2000$ iterations or the pressure array becomes unstable (defined by separation of particles in the pressure mat, e.g. limb poking into mat), the simulation is terminated and the particular $\boldsymbol{\Theta}_{C}$ is rejected. Across the whole dataset, we found roughly a $10 \%$ rejection rate for both of these criteria.

\textbf{Settling criteria - Physics simulation \#2}. We use the same approach as simulation \#1 to determine the height to drop particlized humans. We found it to always be stable for our purpose, and it took roughly 150 iterations to reach the same resting velocity and acceleration previously stated. Because it only uses FleX and the limbs do not move kinematically, it is an order of magnitude faster to run and provides greater flexibility to determine settling criteria. We ran simulation \#2 for a minimum of $200$ iterations and terminated it once the velocity and acceleration thresholds of the particlized human, $v_{ptcl} < 0.05 m /s$ and $a_{ptcl}< 0.5 m / s^2$, were reached. In almost all cases, $200$ iterations was sufficient.



\textbf{Computation time}. For both physics simulations, we ran 10 parallel simulation environments on a computer with 32 cores and a NVIDIA 1070-Ti GPU. This allowed us to generate roughly 35,000 labeled synthetic pressure images per day. 

\begin{figure}
\begin{center}
\includegraphics[width=8.2cm]{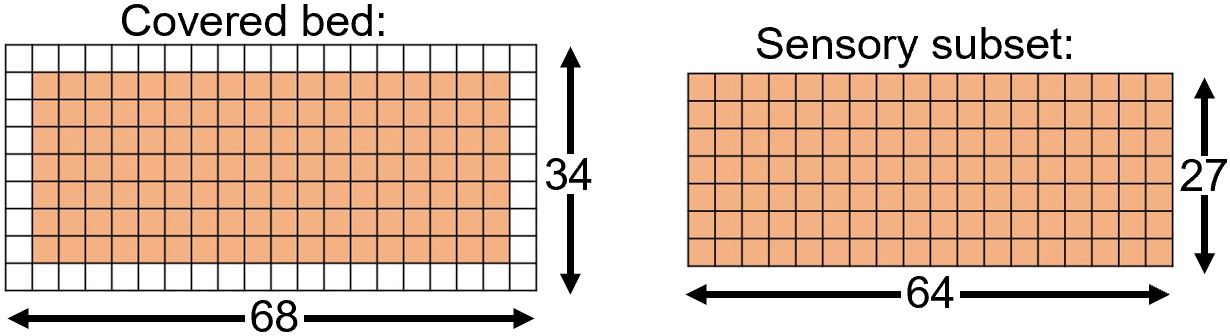}
\vspace{-3mm}
\end{center}
\caption{Size of synthetic pressure mat. Physics simulation \#1 uses forces from particles on the entire covered bed. The pressure mat calculated in physics simulation \#2 uses a smaller subset representing the size of the real pressure mat.}
\label{fig:sensory_subset}
\end{figure}

\begin{figure}
\begin{center}
\includegraphics[width=8.2cm]{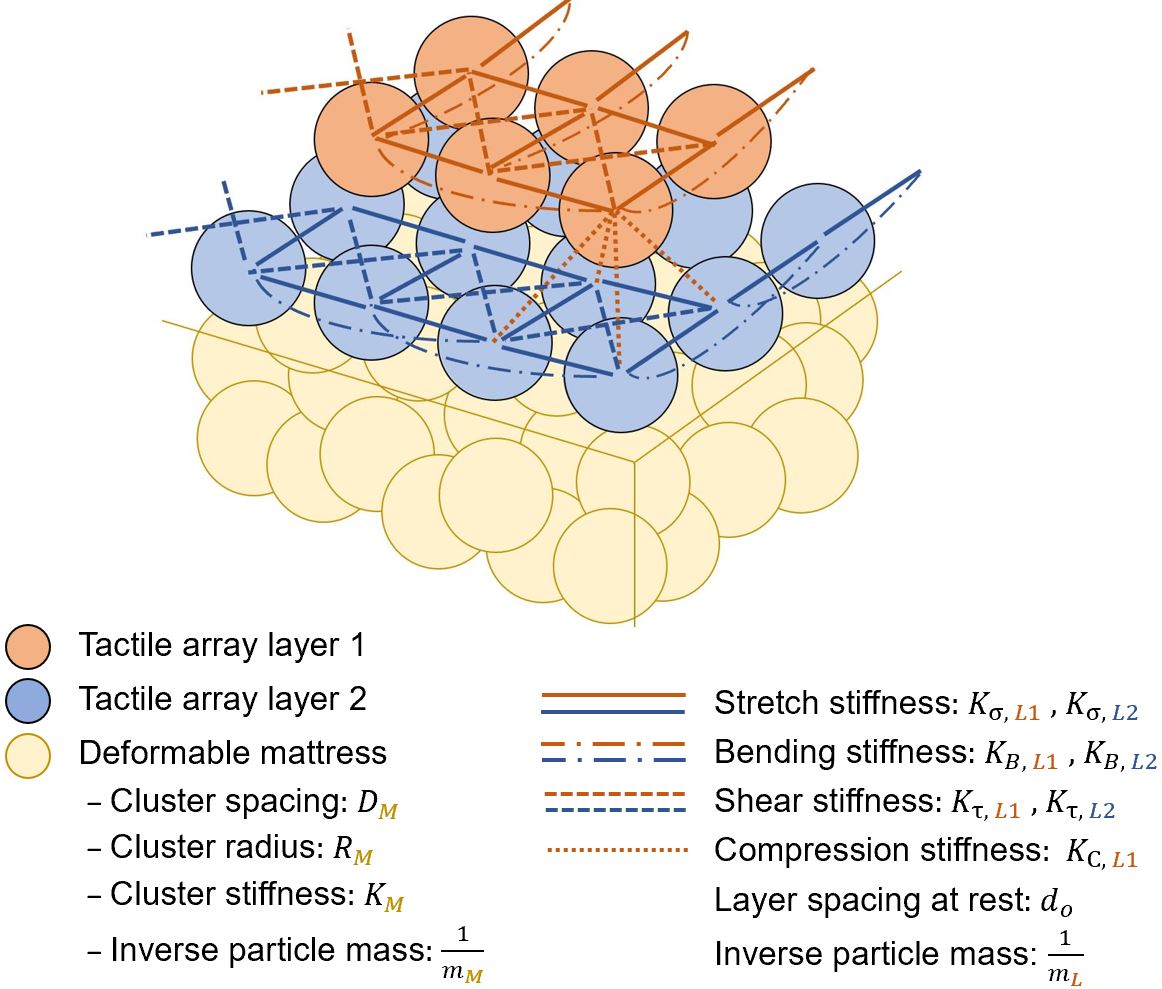}
\end{center}
\vspace{-3mm}
\caption{Pressure mat pyramidal structure showing FleX parameters that we optimized using CMA-ES.}
\label{fig:pmat_structure}
\end{figure}

\subsection{Pressure Mat Structure Details}
\textbf{Limited pressure sensing area.} The sensing portion of the real pressure mat does not cover the entire mattress. We measured a non-sensing border of 6 cm on the sides of the bed and 9 cm at the top and bottom. We built the simulator in the same way: the synthetic pressure mat covers the entire bed (68 x 33), but only an inner subset (64 x 27) representing the sensing area of the pressure image array is recorded, as depicted in Fig. \ref{fig:sensory_subset}.

\textbf{FleX spring constraints.} FleX particles in the synthetic pressure mat are bound together by stiffnesses shown in Fig. \ref{fig:pmat_structure}.

\textbf{Pressure mat adhesion.} For the real pressure mat, velcro and tape are used to prevent sliding across the bed. For the synthetic pressure mat, particles are fixed in horizontal directions across the bed.

\subsection{FleX Calibration}\label{ssec:flexcal}

Although FleX is able to simulate soft bodies, FleX is not optimized to model real-world physics or to calculate realistic pressures. To optimize our FleX simulation to match the real-world mattress and pressure mat, we place a set of static objects on the real mattress, and record the resulting pressure images from the pressure mat. We then build a similar environment in FleX, and we optimize FleX parameters such that the simulated and real-world measurements closely align. 

We jointly optimize 16 deformable bed and pressure sensing array parameters $\mathcal{S}$ using CMA-ES~\cite{hansen2003reducing}. These include the 13 FleX parameters in Fig.~\ref{fig:pmat_structure}, including 4 soft mattress parameters, 7 pressure array stiffnesses, spacing between the pressure mat layers and particle inverse mass, as well as quadratic taxel force constants $C_1$, $C_2$, and $C_3$. To optimize, we first place a set of real rigid objects $\{\mathbbm{o}_1,\ldots \mathbbm{o}_J \}$ each with weights $\{w_1,\ldots w_M\}$ on the real bed. Fig. 5 (a) depicts $\{\mathbbm{o}_1,\ldots \mathbbm{o}_J \}$, where $J=4$ and we use capsular objects with 5 weights for each: 1.3, 2.3, 4.5, 9.1 and 14 kg on the shorter capsules (L=20 cm), and 1.3, 4.5, 9.1, 14 and 18 kg on the longer capsules (L=40 cm). We then collect real pressure mat images $\{\mathbbm{P}_{1,1}, \ldots \mathbbm{P}_{J,M} \}$ and measure the distance that the mattress compresses normal to the bed surface in centimeters, $\{\mathbbm{q}_{1,1}, \ldots \mathbbm{q}_{J,M} \}$. 

Next, we build a matching set of simulated capsules $\{o_1,\ldots o_J \}$ in FleX with the same weights, where one of these objects is shown Fig. 5 (b). At each iteration of the optimization, we drop $J$ simulated capsules of each $M$ weights onto the FleX mattress, re-compute the synthetic pressure images, and compare them to the real ones. The loss function for our optimization takes as input simulated and real pressure images and is computed as:

\begin{equation}{
\arg \underset{\mathcal{S}} \min \sum_{o=1}^J\sum_{k=1}^M{\Big( \mathcal{L}^{F}_{k,o} + \mathcal{L}^{C}_{k,o} + \mathcal{L}^{Q}_{k,o} \Big)}
}\end{equation}
with terms for force error in the pressure mat, $\mathcal{L}^{F}_{k,o}$, contact locations on the pressure mat, $\mathcal{L}^{C}_{k,o}$, and amount of mattress compression by the object, $\mathcal{L}^{Q}_{k,o}$. For some real object $\mathbbm{o}$ with weight $k$ resting on a soft bed at depth $\mathbbm{q}$ from the unweighted height of the soft bed, a pressure image $\mathbbm{P}$ measures forces on individual taxels $\{\mathbbm{u}_1 \ldots \mathbbm{u}_T\}$, where contact is a binary vector $\{\mathbbm{c}_1 \ldots \mathbbm{c}_T\}$ indicating which taxels are measuring non-zero forces. The upper limit $T$ is a spatial index indicating the number of taxels on the pressure image. We note that the value of $T$ for these calibration images is roughly equal to a fraction of the pressure mat size, $(64 \times 27) / 5$, because we drop multiple objects simultaneously to speed up the optimization. Similar to the real mat, the values for the simulated environment are computed as $u_i$, $c_i$, and $q$. The loss terms are computed as:

\vspace{2mm}

\begin{equation}{
\mathcal{L}^{F}_{k,o} ={1\over 2}{\sum_{i=1}^T{|u_i - \mathbbm{u}_i|}\over\sum_{i=1}^T{(u_i+\mathbbm{u}_i )}}+{1\over 2}{|\sum_{i=1}^T{(u_i-\mathbbm{u}_i)}|\over\sum_{i=1}^T{(u_i+\mathbbm{u}_i )}}
}\end{equation}
\vspace{4mm}
\begin{equation}{
\mathcal{L}^{C}_{k,o}={1\over 2}{\sum_{i=1}^T{|c_i - \mathbbm{c}_i|}\over\sum_{i=1}^T{(c_i+\mathbbm{c}_i )}}+{1\over 2}{|\sum_{i=1}^T{(c_i-\mathbbm{c}_i)}|\over\sum_{i=1}^T{(c_i+\mathbbm{c}_i )}}
}\end{equation}




\vspace{1mm}
\begin{equation}{
\mathcal{L}^{Q}_{k,o} = {|q - \mathbbm{q}| \over |q| + |\mathbbm{q}|}
}\end{equation}
The first term for both $\mathcal{L}^{F}_{k,o}$ and $\mathcal{L}^{C}_{k,o}$ account for errors in pressure measurements between individual taxels between the real and simulated pressure mats. The second term accounts for errors in the total measured pressure under an object. All terms are normalized. Since the distances $q$ and $\mathbbm{q}$ are signed, we take the absolute value in the denominator of $\mathcal{L}^{Q}_{k,o}$ for normalization.


\textbf{CMA-ES implementation.} To optimize the FleX environment with CMA-ES \cite{hansen2003reducing}, we used a population size of $50$, max iterations of $3000$, max function evaluations of $1e+8$, mean learning rate of $0.25$, function tolerance of $1e-3$, function history tolerance of $1e-12$, x-change tolerance of $5e-4$, max standard deviation of $4.0$, and stagnation tolerance of $100$. We used a machine with 8 cores and a Nvidia 1070-Ti GPU, and the optimization took 6 days.

Various combinations of parameters result in simulation instability. We perform a constrained optimization by placing a high cost on the evaluation function, \texttt{f\_eval}, when a parameter is suspected of causing instability.  
\begin{itemize}[noitemsep,nolistsep]
    \item Negative FleX parameters can cause instability. If any negative FleX parameter is proposed, a high \texttt{f\_eval} is assigned.
    \item Large differences between $K_{\sigma}, K_{B}, K_{\tau}$ (see Fig. \ref{fig:pmat_structure}) causes knotting in the simulated array. If any stretch, bending, or shear stiffness value is outside of the range $0.5 < K < 2.5$, we add $10x$ the deviation from this range to the \texttt{f\_eval}. 
    \item An unusually long simulation time step indicates instability in the parameters. In this event, the particular rollout is terminated and a high \texttt{f\_eval} is assigned.
    \item If an object takes too long to settle, the rollout is terminated and a high \texttt{f\_eval} is assigned.
\end{itemize}

\subsection{DartFleX Calibration}
The purpose of this calibration is to calibrate the force that should be applied to a DART capsule from particle penetration on the FleX pressure mat. This enables the two simulators to be connected through a mass-spring-damper model, which we described in Section 3.2 in the main paper.

We begin with an optimized FleX environment (Appendix~\ref{ssec:flexcal}) and calibrate the spring coefficient $k$, from the mass-spring-damper model. We calibrate $k$ so that the \textit{dynamic} collision geometries displace the FleX mattress in the same way that real objects would. We take the same set of real objects from the FleX calibration of various shapes $\{\mathbbm{o}_1,\ldots \mathbbm{o}_D \}$ and weights $\{w_1,\ldots w_D\}$, where $D = 20$, place them on the real mattress, and measure the mattress displacement $\{\mathbbm{q}_1, \ldots \mathbbm{q}_D\}$. Then, we recreate the objects as collision geometries $\{\Tilde{o}_1, \ldots \Tilde{o}_D\}$ in FleX, displace the FleX mattress by $\{\Tilde{q}_1,\ldots \Tilde{q}_D, \} = \{\mathbbm{q}_1, \ldots \mathbbm{q}_D\}$, and record the sum of particle penetration distances of underlying taxels $\big\{\sum_{i=1}^P\boldsymbol{x}_{i,1}, \ldots \sum_{i=1}^P\boldsymbol{x}_{i,D}
 \big\}$. We compute $k$ as the average $k$ across $D$ objects:

\vspace{3mm}
\begin{equation}{
k = \Bigg( {w_1 \over \sum_{i=1}^P\boldsymbol{x}_{i,1}}\Bigg|_{\tilde{q}_1} + \ldots + {w_D \over \sum_{i=1}^P\boldsymbol{x}_{i,D}}\Bigg|_{\tilde{q}_D} \Bigg)/D
}\end{equation}
where the vertical bar indicates the amount that object $\Tilde{o}$ of weight $w$ is displaced by distance $\Tilde{q}$, which results in particle penetration distances $\sum_{i=1}^P\boldsymbol{x}_{i}$. The length of a timestep is uncontrollable in FleX. Thus, the timestep in DART is calculated by dropping objects in both environments from a matching height and equating the time to contact the ground, where both simulators have $g=9.81 m/s^2$. This resulted in a DART timestep of $0.0103 s$. 

\subsection{Real Dataset Collection Details}
Participants donned an Optitrak motion capture suit with high contrast to the bed sheets to facilitate analysis of the pose and body shape. We provided S, M, L and XL sizes, and instructed participants to use a form fitting size.

We used the IAI Kinect2 package to calibrate the Kinect~\cite{iai_kinect2}. Our released dataset consists of RGB images and depth/point cloud data from the Kinect that are synchronized and spatially co-registered to the pressure images. We manually synchronized the modalities; only static poses are captured so the time discrepancy is insignificant. We spatially co-registered the Kinect to the pressure mat by putting 1" tungsten cubes on the corners of the pressure mat, which could be seen with all modalities. We captured a co-registration snapshot for each participant, which was taken after they were finished. We created an interface to click on the tungsten block locations on the images and used CMA-ES to find the 6DOF camera pose and co-register it with the mat.

\subsection{Dataset Partitions}

\begin{table}[t!]
\begin{center}
\footnotesize

\renewcommand{\arraystretch}{1.1}
\vspace{1mm}

\scalebox{0.9}{\begin{tabular} {c|c|c|c|c|c}
\hline

\textbf{pose partition}, limb distribution&
\parbox[t]{1mm}{{\rotatebox[origin=c]{90}{gender}}}&  
\parbox[t]{2mm}{{\rotatebox[origin=c]{90}{limbs}}}
\parbox[t]{1mm}{{\rotatebox[origin=c]{90}{on bed}}}&   
\parbox[t]{2mm}{{\rotatebox[origin=c]{90}{train ct.}}}
\parbox[t]{1mm}{{\rotatebox[origin=c]{90}{ synth}}}&  
\parbox[t]{2mm}{{\rotatebox[origin=c]{90}{test ct.}}}
\parbox[t]{1mm}{{\rotatebox[origin=c]{90}{ synth}}}&  
\parbox[t]{2mm}{{\rotatebox[origin=c]{90}{test ct.}}}
\parbox[t]{1mm}{{\rotatebox[origin=c]{90}{ real}}}\\
\hline
\hline

\textbf{general}*   & F & N & 26000 & 3000 & 120\\
even leg space: $\{\mathcal{Y}_1,...\mathcal{Y}_4\} \in \bm{\mathcal{Y}}_L$   & M & N & 26000 & 3000 & 119\\
even arm space: $\{\mathcal{Y}_1, ... \mathcal{Y}_8\} \in \bm{\mathcal{Y}}_A$   & F & Y & 26000 & 3000 & 120\\
  & M & Y & 26000 & 3000 & 120\\
\hline

\textbf{supine general}**  & F & N & 13000 & 1500 & 40\\
even leg space: $\{\mathcal{Y}_1, ... \mathcal{Y}_4\} \in \bm{\mathcal{Y}}_L$   & M & N & 13000  & 1500 & 39\\
even arm space: $\{\mathcal{Y}_1, ... \mathcal{Y}_8\} \in \bm{\mathcal{Y}}_A$    & F & Y & 13000 & 1500 & 40\\
   & M & Y & 13000 & 1500 & 40\\
\hline

\textbf{supine hands behind head}**  & F & Y & 2000 & 500 & 40 \\
even leg space, arms Fig.~\ref{fig:rejection_sampling}(b) & M & Y & 2000 & 500 & 40 \\
\hline

\textbf{prone hands up}$^\dagger$ & F & Y & 4000 & 500 & 40 \\
even leg space, hnds above shldrs & M & Y & 4000 & 500 & 40 \\
\hline

\textbf{supine crossed legs}**  & F & N & 2000 & - & - \\
even leg space, even arm space, & M & N & 2000 & - & - \\
feet must cross according to & F & Y & 2000 & 500 & 40 \\
$x$ direction in Fig.~\ref{fig:rejection_sampling}(a)  & M & Y & 2000 & 500 & 38 \\
\hline
\textbf{supine straight limbs}**   & F & N & 2000 & - & - \\
even leg space, even arm space, & M & N & 2000 & - & - \\
elbows and knees straight  & F & Y & 2000 & 500 & 40 \\
  & M & Y & 2000 & 500 & 36 \\
\hline
\hline
TOTAL & - & - & 184000 & 22000 & 952 \\
\hline
\end{tabular}}
\end{center}
\vspace{-0.2cm}
\caption{Partitions for synthetic data and prescribed poses. For evening the leg space, see  Fig.~\ref{fig:rejection_sampling}(a). For evening the arm space, an additional four subspaces $\{\mathcal{Y}_5, \ldots\mathcal{Y}_8 \}$ are chosen because the most distal joint (hand) is allowed to extend all the way below and above the limb root joint (shoulder), measured in the $y$ direction.\newline
 {\scriptsize * $\theta_{r,3}\sim \mathcal{U}[-{\pi\over 3}, {\pi\over 3}]$, 
 $\theta_{r,1}\sim \mathcal{U}[-\pi, \pi]$
 \newline 
 ** $\theta_{r,3}\sim \mathcal{U}[-{\pi\over 3}, {\pi\over 3}]$, 
 $\theta_{r,1} = 0$
 \newline 
  $^\dagger \hspace{1mm} \theta_{r,3}\sim \mathcal{U}[-{\pi\over 3}, {\pi\over 3}]$, 
 $\theta_{r,1} = \pi$} }
\label{tab:partition_descrip}
\vspace{-0.3cm}
\end{table}

Table~\ref{tab:partition_descrip} presents a detailed description of the data partitions. We split the data for gender. We also split for requiring initial limb positions to be over the surface of the bed, meaning that the Cartesian cuboids used for initial pose sampling (recall Fig.~\ref{fig:rejection_sampling}) are clipped in the $x$ and $y$ directions at the edge of the mattress.

\subsection{Dataset Limitations} 

\textbf{Domain gap.} The real pressure mat has a larger force range. Additionally, as a result of putting a blanket on the bed during the real study, the overall pressure magnitude was reduced $\sim 3x$, which was not reflected in synthetic data calibration. To correct for this, we normalize as described in Appendix~\ref{sec:net_architecture}.

\begin{figure}[t]
\vspace{0.5cm}
\begin{center}
   \includegraphics[width=1.0\linewidth]{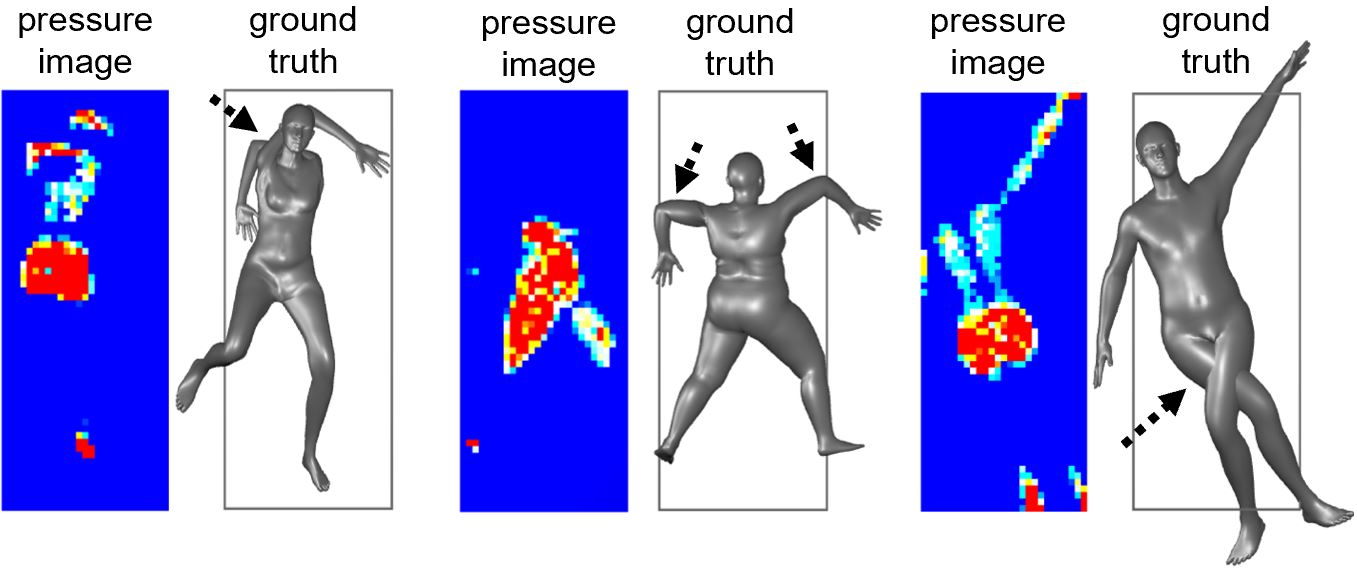}
\end{center}
\vspace{-0.5cm}
   \caption{Uncomfortable or infeasible poses outside of typical human movement range (left, middle). Impossible pose where the thighs are in collision (right).}
\label{fig:pose_discomfort}
\vspace{-0.5cm}
\end{figure}

\begin{figure*}
\begin{center}
\includegraphics[width=17.5cm]{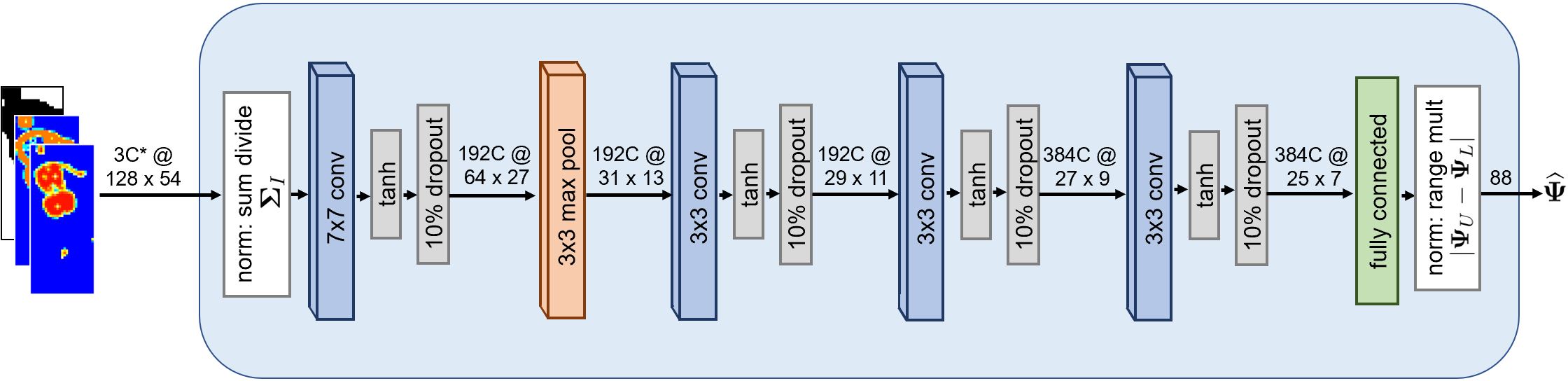}
\end{center}
\vspace{-3mm}
    \caption{PressureNet: Convolutional Neural Network (CNN) with five convolutional layers, one max pooling layer, and one fully connected layer. Input images are normalized by per-image division by the sum of taxels. * indicates that the number of channels shown (3) represents Mod1 in Fig. 6 (a), whereas Mod2 in Fig. 6 (a) uses 5 input channels.}
\label{fig:cnn_block}
\end{figure*}

\textbf{Synthetic body joint limits.} We observed that roughly $2\%$ of the synthetic poses appear uncomfortable or infeasible for a real person (Fig~\ref{fig:pose_discomfort}). This work could be improved by using pose-conditioned joint angle limits such as \cite{akhter2015poseSHORT} instead of constant limits. Fig.~\ref{fig:pose_discomfort}-right shows an impossible pose where the thighs are in collision. We were not able to check collisions between the thighs using the capsulized model because the thigh capsules are often in collision for valid poses.

\begin{figure*}
\begin{center}
\includegraphics[width=17.5cm]{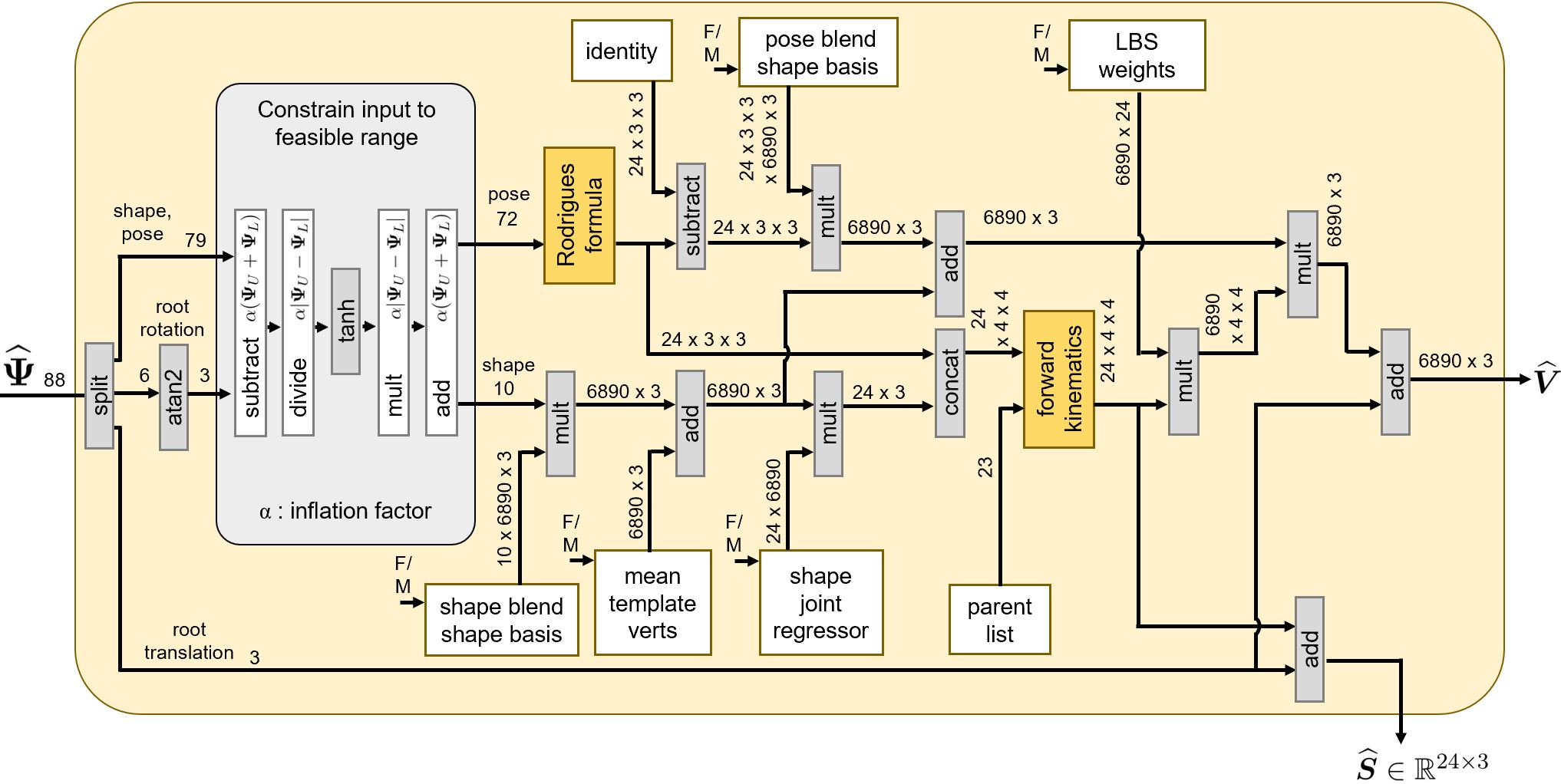}
\end{center}
\vspace{-3mm}
    \caption{PressureNet: Differentiable SMPL human mesh reconstruction from Kanazawa et al. \cite{kanazawa2018end}. Our additions to \cite{kanazawa2018end} include input constraints (shown in the light grey box) and the root joint rotation and translation.
    }
\label{fig:hmr_block}
\end{figure*}

\begin{figure*}
\begin{center}
\includegraphics[width=17.5cm]{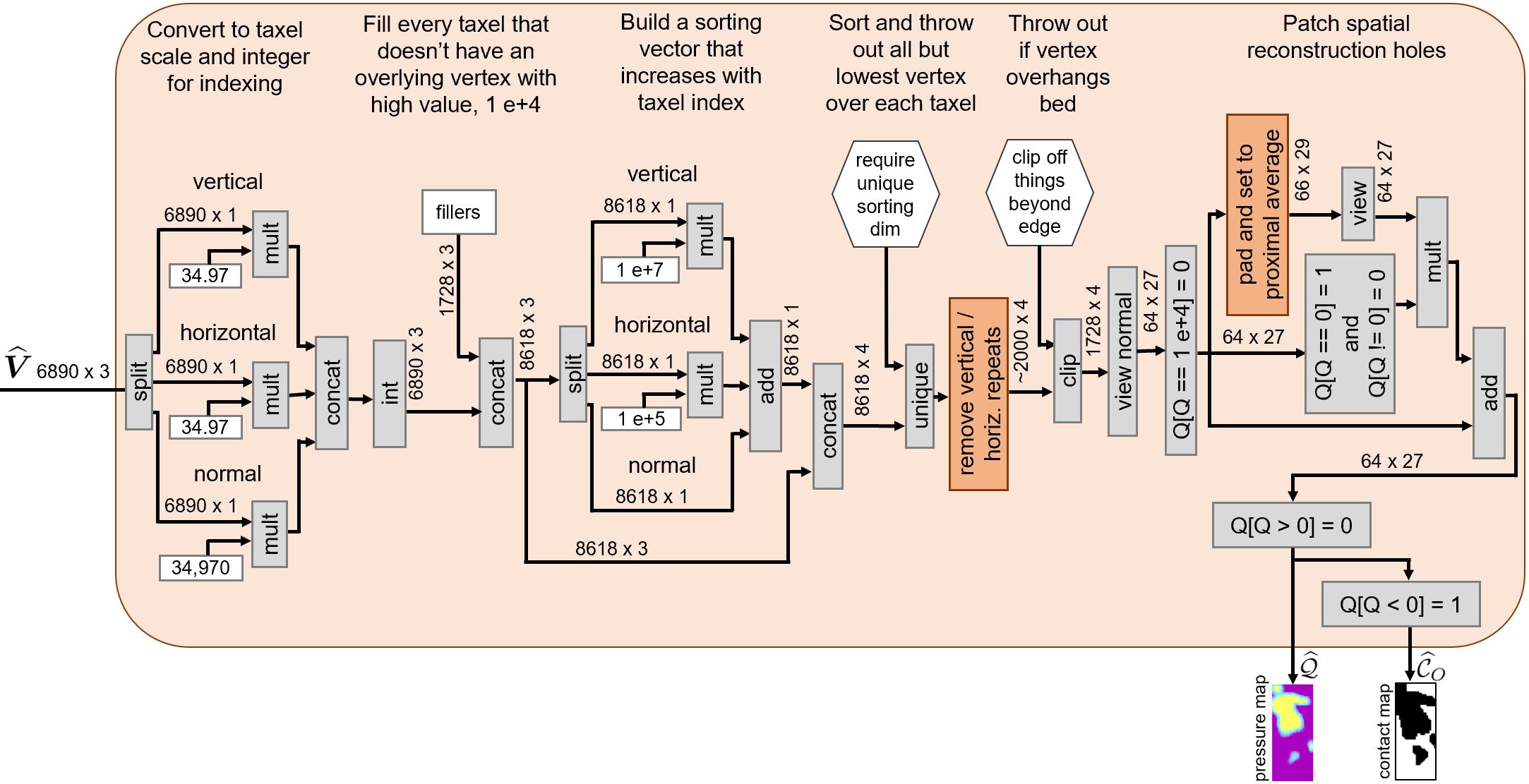}
\end{center}
\vspace{-3mm}
    \caption{PressureNet: Pressure Map Reconstruction (PMR). PMR is fully differentiable, and performs sorting, filtering and patching to reconstruct spatial maps from the human mesh.}
\vspace{-3mm}
\label{fig:pmr_block}
\end{figure*}

\section*{Appendix B: PressureNet}

\setcounter{subsection}{0}  

\subsection{PressureNet Architecture Details}\label{sec:net_architecture}
\textbf{CNN - Convolutional Neural Network}. Our CNN architecture, depicted in Fig.~\ref{fig:cnn_block}, is similar to that of Clever et al. \cite{clever20183d}, and uses the same kernel sizes, layers, and dropout. The first layer is a convolutional layer with a 7x7 kernel, and uses a stride of 2 and zero padding of size 3 on the sides of input images. 
The max pooling layer has a stride of 2 and padding of 0. All other convolutional layers are 3x3 with a stride of 1 and padding of 0. We use 192 channels in the first two convolutional layers and in the max pooling layer, and 384 channels in the last two convolutional layers. This CNN also differs from \cite{clever20183d} in that we use tanh activation functions instead of ReLU. Through informal testing on smaller data sizes (e.g. 46K images), we observed that networks with tanh activations had less overfitting.  We normalize the input and output of the network. To normalize the input channels, divide by the sum of taxels for each input image, $\bm{\Sigma}_I$. To normalize the output, we multiply it by the range of shape, pose, and posture parameters from the synthetic training dataset. We compute the range from the lower and upper limits,$\boldsymbol{\Psi}_L$ and $\boldsymbol{\Psi}_U$, of all parameters in the training dataset. For joint angle limits (i.e. pose), we use values from \cite{soucie2011range, boone1979normal, roaas1982normal}. For body shape, we use sampling bounds $[-3, 3]$ from \cite{ranjan2018learning}. For global rotation, we use our sampling bounds for roll and yaw of $[-\pi, \pi]$ and $[-{\pi\over 6}, {\pi \over 6}]$, and for global translation, we use the size of the bed. 

\textbf{SMPL - Human Mesh Reconstruction}. Following the CNN, we use the human model generative part of the HMR network \cite{kanazawa2018end}, which inputs estimated shape, pose, and posture $\boldsymbol{\hat{\Psi}}$, and outputs a differentiable human mesh reconstruction $\boldsymbol{\hat{V}}$, as well as a set of $N=24$ Cartesian joint positions $\boldsymbol{\hat{S}}$. This generative SMPL model, implemented in PyTorch~\cite{mo2018hmr}, along with our modifications, is presented in Fig.~\ref{fig:hmr_block}.

In addition to using the generative kinematic SMPL embedding part of the full HMR network, our implementation constrains the input parameters to keep angles within human limits and body shape parameters inside our initial sampling range. To constrain the input parameters, we normalize the parameters to a range $[-1, 1]$ based on the limits $\boldsymbol{\Psi}_L$, $\boldsymbol{\Psi}_U$, and use a tanh function for a soft limit that is more amenable to gradient descent. Then, we perform a reverse normalization to scale back up. To prevent the tanh from clipping feasible values at the angle limits, for example a straight knee that is at 0 degrees, we inflate the angle range by a factor $\alpha = 1.2$ as shown in the figure.

\textbf{PMR - Pressure Map Reconstruction}. 
PMR, a novel component of PressureNet, takes as input a human mesh in global space $\boldsymbol{\hat{V}}$, and outputs a set of reconstructed spatial maps $ \{ \widehat{\mathcal{Q}},  \widehat{\mathcal{C}}_O \}$, which resemble a real pressure image and indicate where contact occurs between the estimated mesh and the bed. 
We reconstruct these maps differentiably as depicted in Fig. \ref{fig:pmr_block}, meaning that we can backpropagate gradients through PMR to train the CNN. The PMR loss is based on the error between estimated spatial maps $ \{ \widehat{\mathcal{Q}},  \widehat{\mathcal{C}}_O \}$ and ground truth spatial maps $ \{\mathcal{Q},  \mathcal{C}_O \}$.  PMR works by projecting the mesh onto the surface of the bed and computing the distance that it sinks into the bed over each taxel. This amounts to finding the distance between the lowest vertex within the $2.9 \times 2.9$ cm area of each taxel and the undeformed height of the bed.

The PMR input $\boldsymbol{\hat{V}}$ is in units of meters, which we convert to units of taxels ($ 1 \text{ m} \sim 35 \text{ taxels}$), so it can be indexed on the scale of the pressure image. We then use a process involving sorting, filtering, and patching to recreate the spatial maps, which is detailed in Fig.~\ref{fig:pmr_block}.

\begin{figure}
\begin{center}
\includegraphics[width=7.5cm]{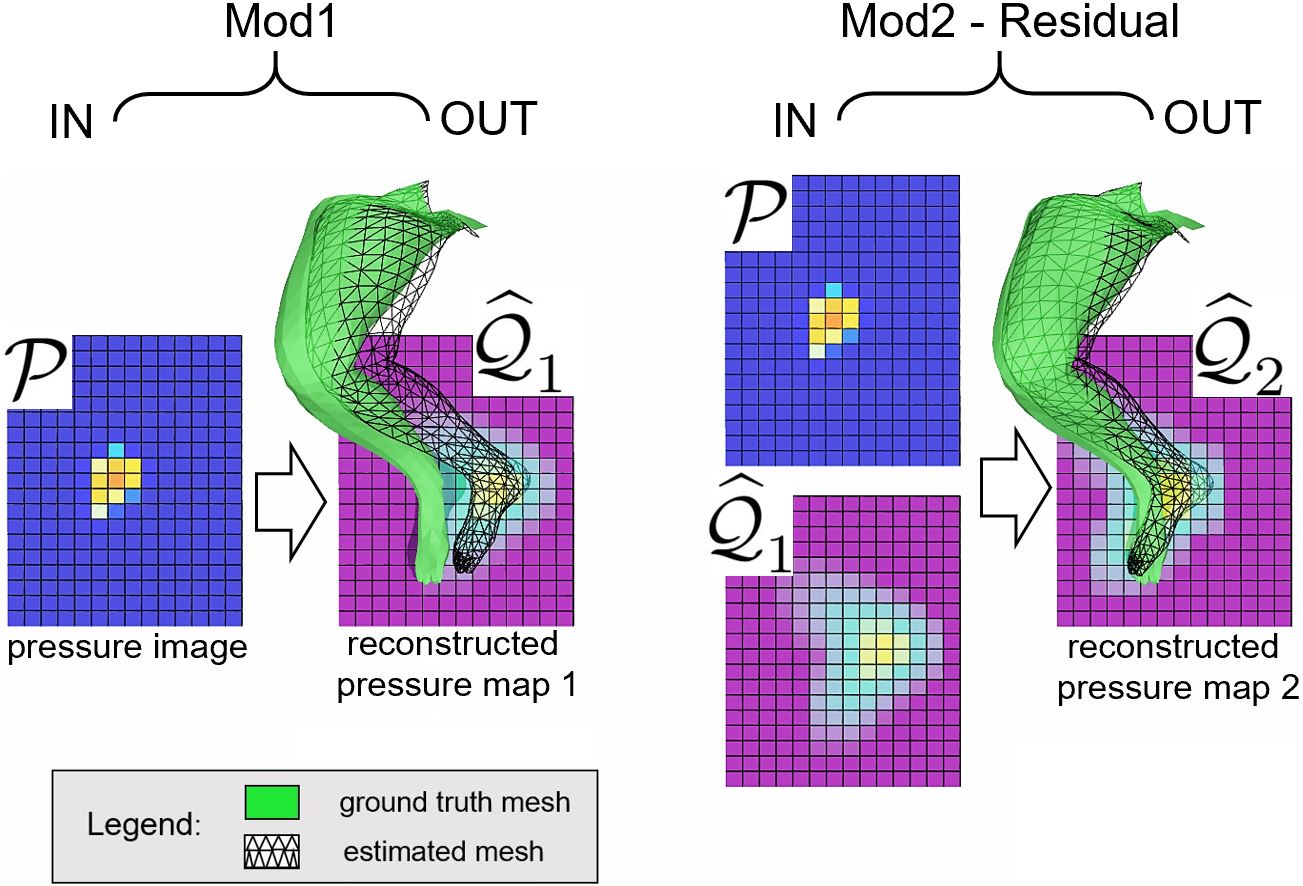}
\end{center}
    \caption{PressureNet deep learning in action, showing an example from our synthetic test set. The first network module (``Mod1'') outputs an initial coarse pose estimate (right leg shown) and a reconstructed pressure map $\widehat{\mathcal{Q}}_1$. The second network module (``Mod2'') corrects the estimated black mesh by a small angle difference based on the spatial residual between $\mathcal{P}$ and $\widehat{\mathcal{Q}}_1$.}
\vspace{-3mm}
\label{fig:contact_aware}
\end{figure}

\subsection{PressureNet Loss Function}

We compute a loss on joint error rather than vertex error because the vertices are highly concentrated in some areas like the face and hands for aesthetic reasons, rather than for representing overall pose. Moreover, training the first network module (``Mod1'') with reconstruction of 24 joint positions rather than a full set of vertices is much faster.

The purpose of the second network module (``Mod2'') is to fine-tune an initial estimate from Mod1 using both reconstructed pressure maps as input and a loss function with spatial map awareness. Fig. \ref{fig:contact_aware} shows a real example of how Mod2 corrects the initial mesh estimate from Mod1 using PMR. Note the spatial difference in the input images for Mod2, where the reconstructed map of the foot pressure in $\widehat{\mathcal{Q}}_1$ is shifted further right than the information on pressure image $\mathcal{P}$.

\subsection{PressureNet Training Details}
We build PressureNet in PyTorch \cite{paszke2017automatic}, which is shown at a high level in Figure 6 (b). 
For both Mod1 and Mod2, we used a learning rate of $0.00002$ and a weight decay of $0.0005$, which are the same used in \cite{clever20183d}. We used the Adam optimizer for gradient descent \cite{kingma2014adam}. Training Mod2 for 100 epochs using 184K images took 3 days on a Nvidia Tesla K80 GPU. Training Mod2 took 8 days due to increased computation from PMR. 

\subsection{Results for Separate Partitions}

\begin{table}[t!]
\begin{center}
\footnotesize

\renewcommand{\arraystretch}{1.1}
\vspace{1mm}

\scalebox{0.9}{\begin{tabular} {c|c|c|c|c}
\hline
 & 
test ct. & 
test ct. &  
3DVPE &
3DVPE\\
pose partition & 
real & 
synth &  
real (cm) &  
synth (cm)\\
\hline
\hline

supine straight limbs & 76 & 1000 & 3.71 & 2.68 \\
supine general & 159 & 2000 & 4.51 & 3.40 \\
supine crossed legs & 78 & 1000 &  4.49 & 3.41  \\
prone hands up & 80 & 1000 & 5.12 & 4.24  \\
general, roll $\sim \mathcal{U}[-\pi, \pi]$ & 479 & 6000 & 5.39 & 4.30 \\
supine hands behind head & 80 & 1000 & 5.09 & 4.40 \\
\hline
\cline{1-5}
gender partition  \\
\cline{1-5}
F & 480  & 6000 & 4.88 & 3.85 \\
M & 472 & 6000 & 5.10 & 4.04 \\
\hline
\end{tabular}}
\end{center}
\vspace{-0.2cm}
\caption{\label{tab:results_partition}Partitioned results for prescribed poses with the best network for each real and synthetic. } 
\end{table}

Table~\ref{tab:results_partition} shows the results of our PressureNet evaluated between prescribed resting poses from participants in bed, and a per-gender comparison.

\subsection{Additional Failure Cases} We present additional failure cases in Fig.~\ref{fig:failure_cases_appendix}. One limitation is that our network does not have an interpenetration error, so the limbs sometimes intersect, e.g. the left hand in Fig.~\ref{fig:failure_cases_appendix}(a)-top left. Our network also failed for some limbs when there was little or no contact information, and for non-resting poses. This issue is related to the limitations of the sensor, which were explored in \cite{clever20183d}. Our network failed for non-resting poses, such those in \cite{clever20183d}; however these are not part of the training or testing PressurePose dataset. We observed some inaccuracies when testing on training data (Figs. 9 and~\ref{fig:failure_cases_appendix}), which suggests that there is a performance limitation on the network's ability to extract pressure image features in some scenarios.

\begin{figure}[t]
\begin{center}
   \includegraphics[width=1.0\linewidth]{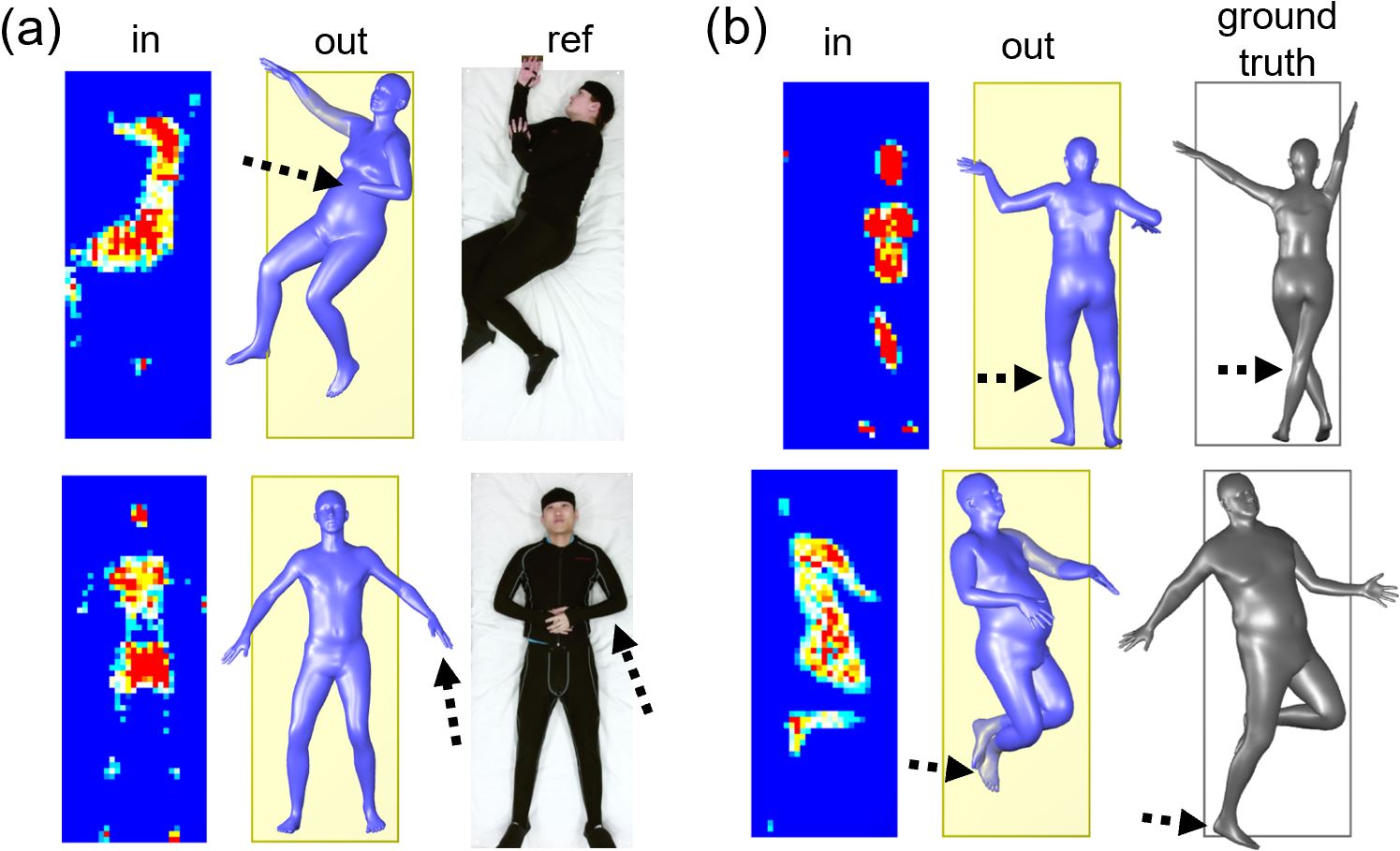}
\end{center}
\vspace{-0.4cm}
   \caption{(a) Real data failure cases. Self penetration of inferred left hand into chest (top), lack of information on mat leading to inaccurate pose (bottom). (b) Synthetic data failure cases: testing on  \textit{training} data, various inaccuracies.}
\label{fig:failure_cases_appendix}
\vspace{-0.5cm}
\end{figure}

\end{alphasection}

{\small
\bibliographystyle{ieee_fullname}
\bibliography{main}
}

\end{document}